\documentclass{article}

\usepackage[final]{corl_2022} 

\usepackage{times}
\usepackage{algorithmic}
\usepackage{algorithm}
\usepackage{amsmath}
\usepackage{amssymb}
\usepackage{amsthm}
\usepackage{graphics}
\usepackage{epsfig}
\usepackage[caption=false, font=footnotesize]{subfig}
\usepackage{multirow}
\usepackage{wrapfig}
\usepackage{mathtools}
\usepackage{makecell}
\usepackage{bbm}
\usepackage{booktabs}
\usepackage{graphicx}
\usepackage{enumitem,kantlipsum}

\usepackage[numbers]{natbib}
\usepackage{multicol}

\newcommand{\SO}{\mathrm{SO}}
\newcommand{\OO}{\mathrm{O}}

\newcommand{\plotDescription}{The plots show the performance of the behavior policy in terms of the discounted reward. Each point is the average discounted reward in the previous 500 steps. Results are averaged over four runs. Shading denotes standard error.}

\newcommand{\edit}[1]{#1}

\title{On-Robot Learning With Equivariant Models}

\author{Dian Wang \quad Mingxi Jia \quad Xupeng Zhu \quad Robin Walters\quad Robert Platt\\
Khoury College of Computer Sciences\\
Northeastern University\\
Boston, MA 02115, USA \\
\texttt{\{wang.dian,jia.ming,zhu.xup,r.walters,r.platt\}@northeastern.edu}
}

\begin{document}
\maketitle


\begin{abstract}
Recently, equivariant neural network models have been shown to improve sample efficiency for tasks in computer vision and reinforcement learning. This paper explores this idea in the context of on-robot policy learning in which a policy must be learned entirely on a physical robotic system without reference to a model, a simulator, or an offline dataset. We focus on applications of Equivariant SAC to robotic manipulation and explore a number of variations of the algorithm. Ultimately, we demonstrate the ability to learn several non-trivial manipulation tasks completely through on-robot experiences in less than an hour or two of wall clock time.  
\end{abstract}

\section{Introduction}

Training directly on a physical robot is challenging because it takes a long time to gather experiences: an environmental step on a physical robot system is often at least one or two orders of magnitude slower than an environmental step in the simulation. As a result, it is not unusual for researchers who want to learn on a physical robotic system to spend hundreds of hours of robot time to learn even simple manipulation control policies, e.g.~\cite{pinto2016supersizing,qt_opt,berscheid2021robot}. However, recent work has shown that policy learning using equivariant models with rotation, translation, and reflection symmetries yield much higher sample efficiency than conventional approaches~\cite{van2020mdp,iclr}. 

This paper demonstrates that with the right choices for symmetry group and data augmentation strategy, these equivariant approaches can be so sample efficient that it becomes feasible to learn simple robotic manipulation policies directly on a physical system (we call this \emph{on-robot} learning). This newfound ability to learn simple manipulation skills quickly and directly on a robot also gives us a new perspective on the problem of the sim2real gap, the small differences between a simulation of the real world and the real world itself~\cite{hofer2020perspectives}. 
Our results suggest that it is not always worthwhile to train a policy in simulation first before fine-tuning in the real world, at least in the context of simple manipulation tasks where equivariant models can learn quickly.

This paper makes three contributions. First, we find that equivariance with respect to discrete symmetry groups leads to better performance than equivariance with respect to continuous groups.
Although prior work~\cite{van2020mdp,iclr} makes it clear that equivariant policy learning can be much more sample efficient than learning with non-equivariant models, it is not clear what symmetry groups are most appropriate in robotic domains with rotation and reflection symmetries.
Based on recent work, we know we can encode continuous $\SO(2)$ and $\OO(2)$ symmetries using the irreducible representations of the group~\cite{e2cnn}. However, approximating continuous symmetry using discrete subgroups may also work well.
This paper evaluates these alternatives and finds that even though the continuous group more closely reflects the actual problem symmetries, discrete rotation and reflection groups like $D_4$ and $C_8$ still have better performance.

Second, we show data augmentation further improves models equivariant to discrete groups. Since equivariant models hard code problem symmetries into the neural network model, one might assume that data augmentation would no longer be helpful. In the case of discrete groups, which only approximate the full domain symmetry, this turns out to be untrue. Here, we evaluate equivariant policy learning with and without various types of data augmentation and find that even something as simple as buffer augmentation much improves performance.


Third, we perform a series of evaluations of on-robot learning using equivariant methods. We show that the equivariant models are so sample efficient that they can learn policies for solving various manipulation tasks from scratch within one or two hours. Furthermore, we demonstrate that sim2real pre-training is unnecessary for equivariant policy learning and is sometimes harmful.
Since equivariant policy learning makes it possible to learn simple manipulation policies directly on a physical robot efficiently, it is worth asking whether the sim2real approach is still useful in these applications. We compare training exclusively on the robot with a sim2real strategy where we pre-train in a PyBullet simulation and then transfer onto the physical robot. We find -- in the four representative manipulation applications explored here -- that while there is often a benefit to training in simulation first before transferring to the robot, this is by no means necessary. Moreover, it is sometimes the case that the simulation and physical agents learn qualitatively different things, potentially leading to negative sim2real transfer, i.e., the situation where the pre-trained policy actually impedes learning on the physical system. Supplementary video and code are available at \url{https://pointw.github.io/equi_robot_page/}.

\section{Related Work}

\textbf{Equivariant Learning}:  The first equivariant neural networks introduced were $G$-Convolution~\cite{g_conv} and Steerable CNNs~\cite{steerable_cnns}, which improved the sample efficiency of traditional convolutional neural networks by injecting symmetries in the structure of the neural network. \citet{e2cnn} proposed a framework for implementing general $\mathrm{E}(2)$-Steerable CNNs. Recent work showed encouraging results for applying equivariant networks in various computer vision~\cite{benton2020learning,dey2020group} and dynamics~\cite{wang2020incorporating, walters2020trajectory} tasks. They have also been applied to deep RL~\cite{mondal2020group,van2020mdp} and robotic manipulation~\cite{corl,iclr,zhu2022grasp,equi_transporter} with compelling results. 
However, to our knowledge, equivariant methods have never been explored in the context of on-robot reinforcement learning.


\textbf{On-Robot Learning}: The most common approach to robotic policy learning is to train in simulation and then transfer to a real world application~\cite{rusu2017sim,zhu2018reinforcement,gualtieri2020learning,wang2020policy,biza2021action}. 
Nevertheless, there have been several efforts to develop methods that enable an agent to learn a policy directly on a physical robotic system. 
\citet{gu2017deep} trained manipulation skills in fixed environments with multiple physical robot workers. \citet{singh2019end} developed a method that learned manipulation skills within 1-4 hours in the real world but required a user to respond to queries for labels. \citet{qt_opt} trained a grasping policy with seven robots and over 800 robot hours. \citet{zeng2018learning, zeng2020tossingbot} 
demonstrated on-robot learning by encoding the $Q$ function using a fully convolutional network, but only in the context of open-loop tasks where the gripper performed a pre-defined behavior.
FERM~\cite{ferm} performed on-robot learning using SAC~\cite{sac} in combination with a contrastive learning objective~\cite{oord2018representation, curl}, but only for tasks where the orientation of the gripper was fixed. 
Relative to the work above, our method is most comparable to FERM~\cite{ferm}, and we therefore benchmark our method against that.


\section{Background}



\textbf{Equivariance Over the Rotation Group:} Many robotics problems display rotational and reflectional symmetry in the plane perpendicular to gravity. These are captured by the group $\OO(2)$ which contains all continuous planar rotations $\mathrm{Rot}_\theta$ about the origin and reflections through lines through the origin. It contains the subgroup of rotations $\SO(2)=\lbrace \mathrm{Rot}_\theta : 0 \leq \theta < 2 \pi \rbrace$. 
Sometimes, we are interested in discrete subsets, for example the cyclic subgroup $C_n = \lbrace \mathrm{Rot}_\theta : \theta \in \lbrace \frac{2\pi i}{n} | 0 \leq i < n \rbrace \rbrace $ or the Dihedral group $D_n$ which contains the $n$ rotations of $C_n$ as well as $n$ reflections through $n$ evenly spaced lines through the origin. 
Domain symmetries can be described as invariance or equivariance of task functions.
A function $f$ is \emph{$G$-invariant} if when its input $x$ is transformed by a symmetry group element $g\in G$, its output stays the same, $f(gx) = f(x)$. A function $f$ is \emph{$G$-equivariant} if when its input $x$ is transformed by a symmetry group element $g\in G$, its output transforms accordingly by $g$, $f(gx) = gf(x)$.

\textbf{Group Invariant MDPs:} Equivariant policy learning uses symmetries in the MDP to structure the neural network model used to represent the policy and value function. Let $M=(S, A, T, R, \gamma)$ denote an MDP and let $g \in G$ denote an element of a symmetry group $G$ (e.g., $G=\OO(2)$). We will say that MDP $M$ is $G$-invariant if both the transition function and the reward function are invariant: $T(s,a,s') = T(gs,ga,gs')$ and $R(s,a) = R(gs,ga)$ for all $g \in G$~\cite{iclr}. This type of symmetry is a good fit for robotics problems that are invariant over rotation and reflection. In many manipulation problems, for example, the objective is to perform some task (e.g., open a drawer or insert a part) regardless of the respective poses of the parts involved.


\begin{wrapfigure}[18]{r}{0.57\textwidth}
\vspace{-0.7cm}
\centering
\subfloat[Equivariant Actor]{\includegraphics[height=0.49\linewidth]{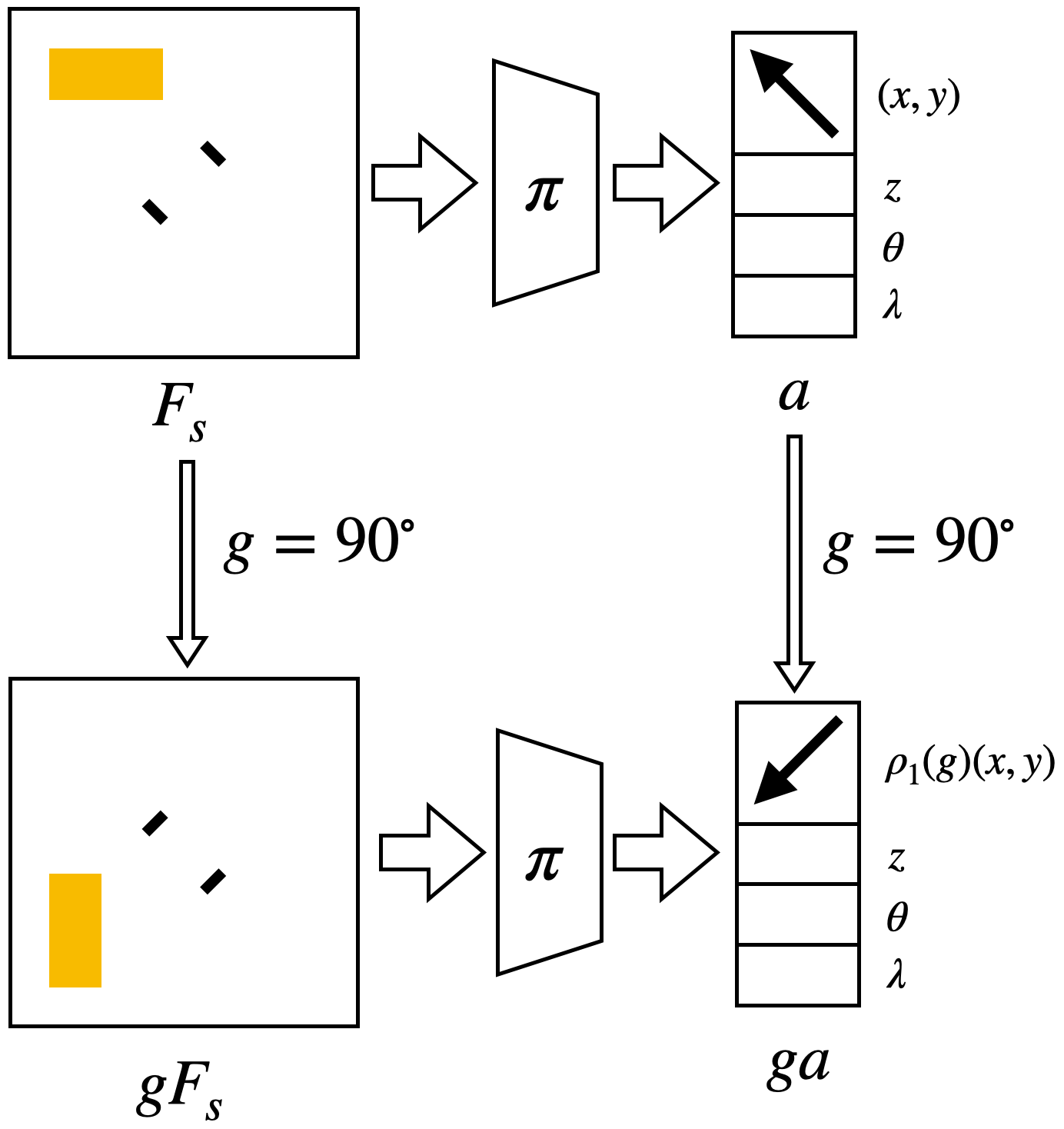}}
\subfloat[Invariant Critic]{\includegraphics[height=0.49\linewidth]{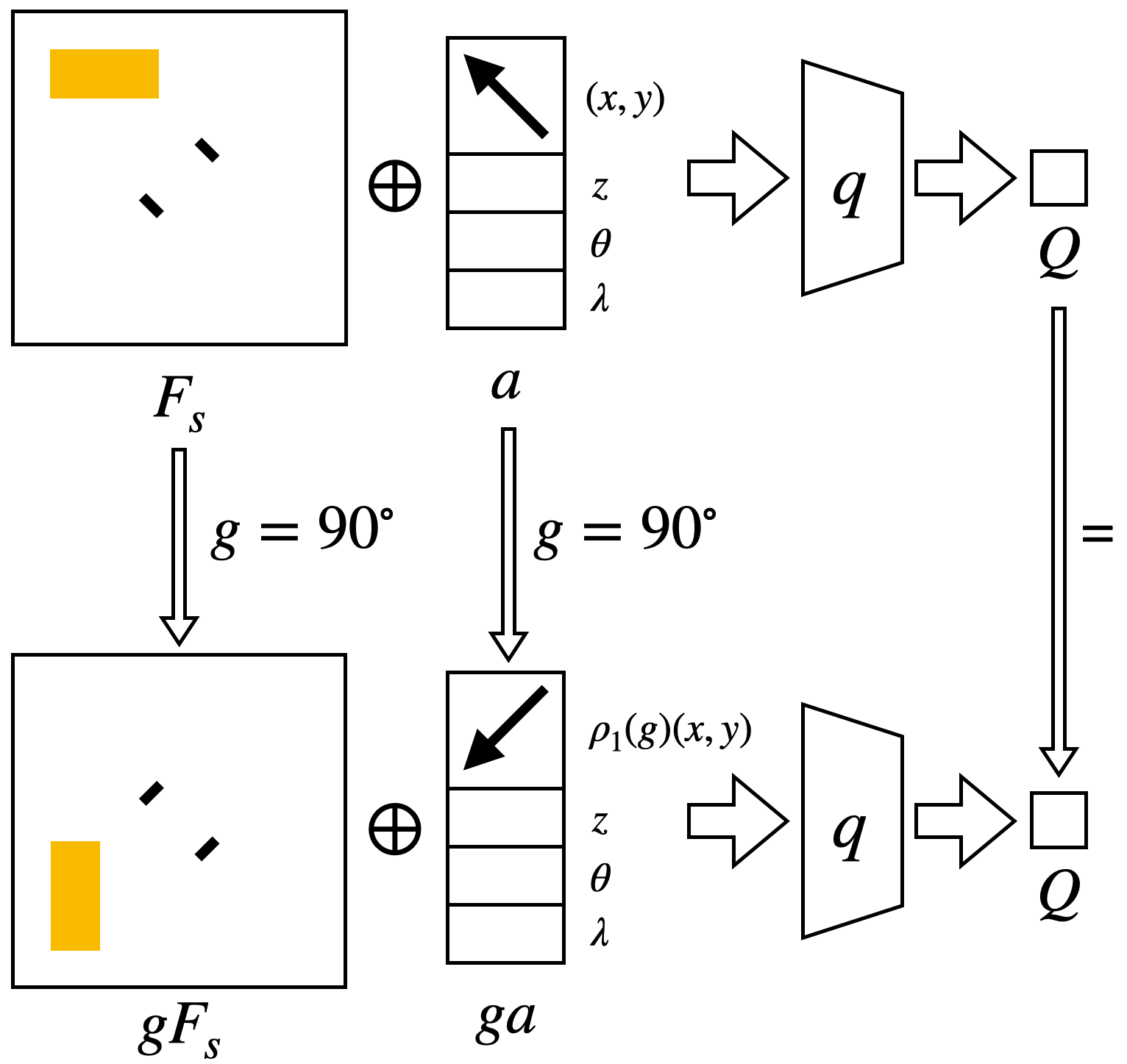}}
\caption{Illustration of the Equivariant SAC. (a): the equivariant actor's output action rotates as the input state rotates. (b): the invariant critic's output doesn't change when the input state and action are rotated simultaneously.}
\label{fig:equi_sac}
\end{wrapfigure}


\textbf{Equivariant SAC:} Equivariant Soft Actor Critic (Equivariant SAC)~\cite{iclr} is a version of SAC~\cite{sac} that uses equivariant neural models to encode the symmetries of a $G$-invariant MDP. It implements the critic using an $G$-invariant model, $q(gs, ga) = q(s, a)$, and the actor using an $G$-equivariant model, $\pi(gs) = g\pi(s)$. This is illustrated in Figure~\ref{fig:equi_sac} for a manipulation domain where $G=\SO(2)$, state is an image $s=\mathcal{F}_s$, and action is a vector $a = (x, y, z, \theta, \lambda)$, where $(x, y, z)$ is the gripper position displacement, $\theta$ is the gripper orientation displacement about the $z$-axis, and $\lambda$ is the gripper aperture. The action $a$ is partitioned into $(x,y) \in A_{equi}$ and $(z, \theta, \lambda) \in A_{inv}$ so that $ga$ rotates the $(x,y)$ component of action but leaves the other action dimensions unchanged. Figure~\ref{fig:equi_sac} left shows the equivariance of the actor. When the state image rotates by $90$ degrees, the $x,y$ components of action rotate but the $z, \theta, \lambda$ components remain unchanged. The right side shows the invariance of the critic. Corresponding rotations of state and action do not change the output $Q$ value.

\section{Symmetry Group and Augmentation Strategy}

\label{sec:sim_exp}

\begin{figure*}[t]
\centering
\newlength{\env}
\setlength{\env}{0.115\textwidth}
\subfloat[Block Picking]{
\label{fig:sim_env_pick}
\includegraphics[width=\env]{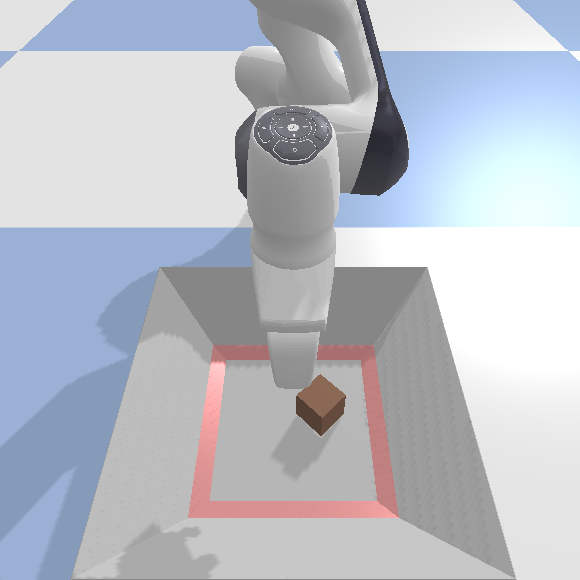}
\includegraphics[width=\env]{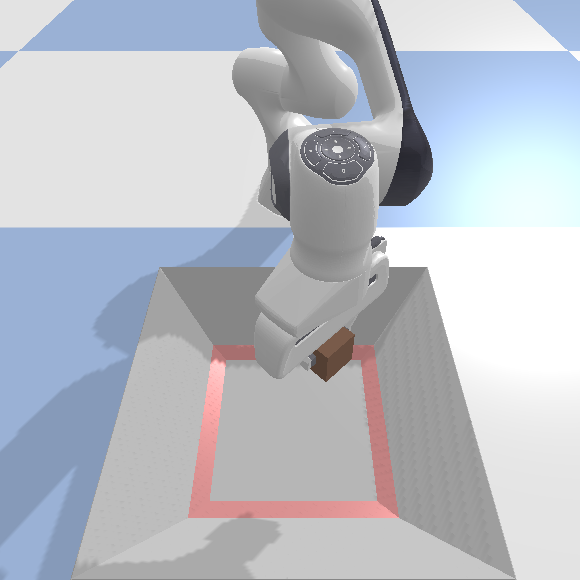}
}
\subfloat[Clutter Grasping]{
\label{fig:sim_env_grasp}
\includegraphics[width=\env]{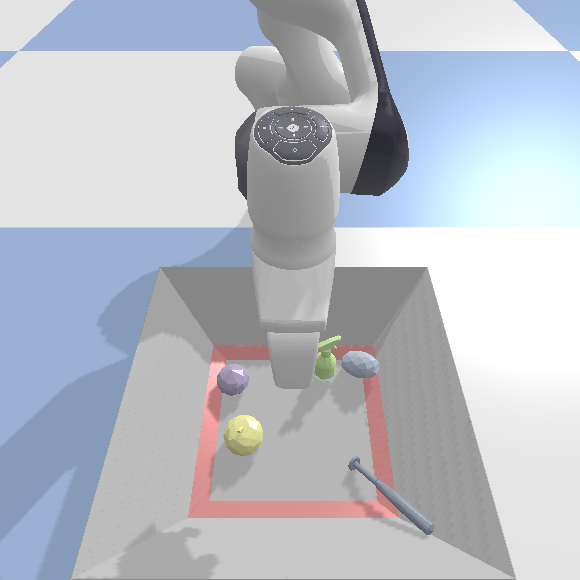}
\includegraphics[width=\env]{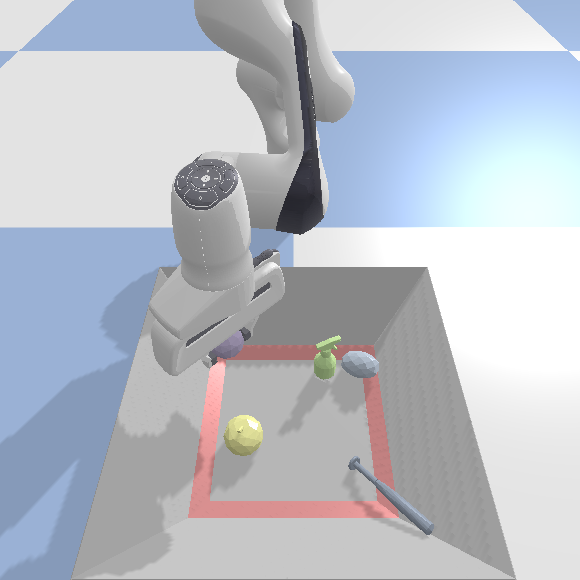}
}
\subfloat[Block Pushing]{
\label{fig:sim_env_push}
\includegraphics[width=\env]{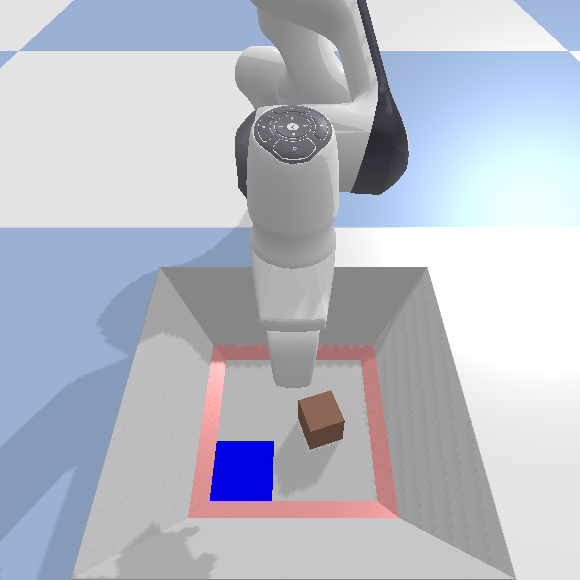}
\includegraphics[width=\env]{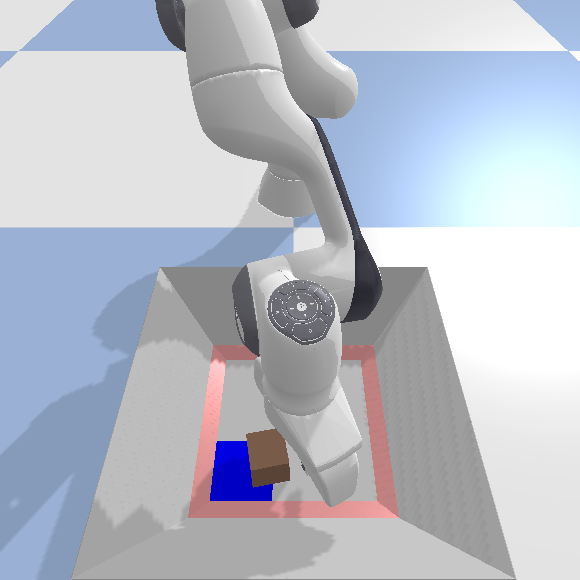}
}
\subfloat[Block in Bowl]{
\label{fig:sim_env_bowl}
\includegraphics[width=\env]{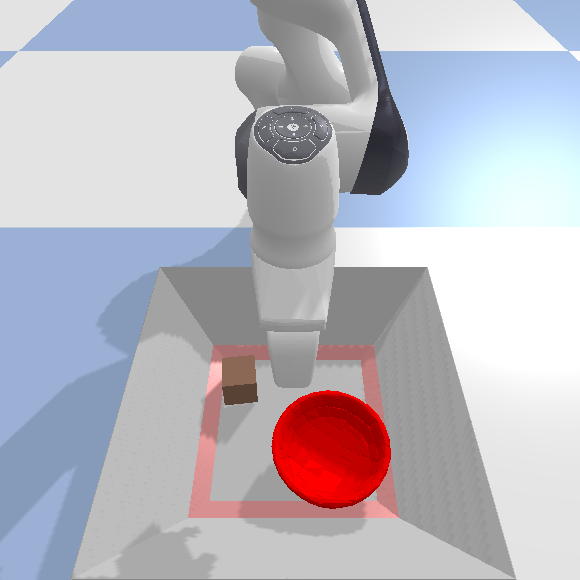}
\includegraphics[width=\env]{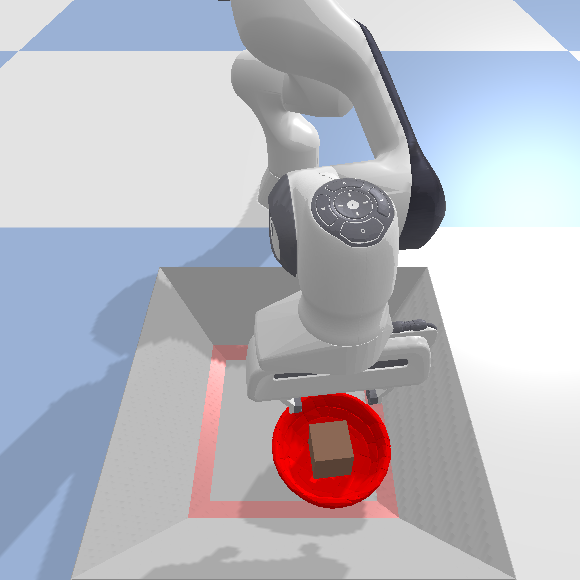}
}\\

\subfloat[Block Picking]{
\includegraphics[width=\env]{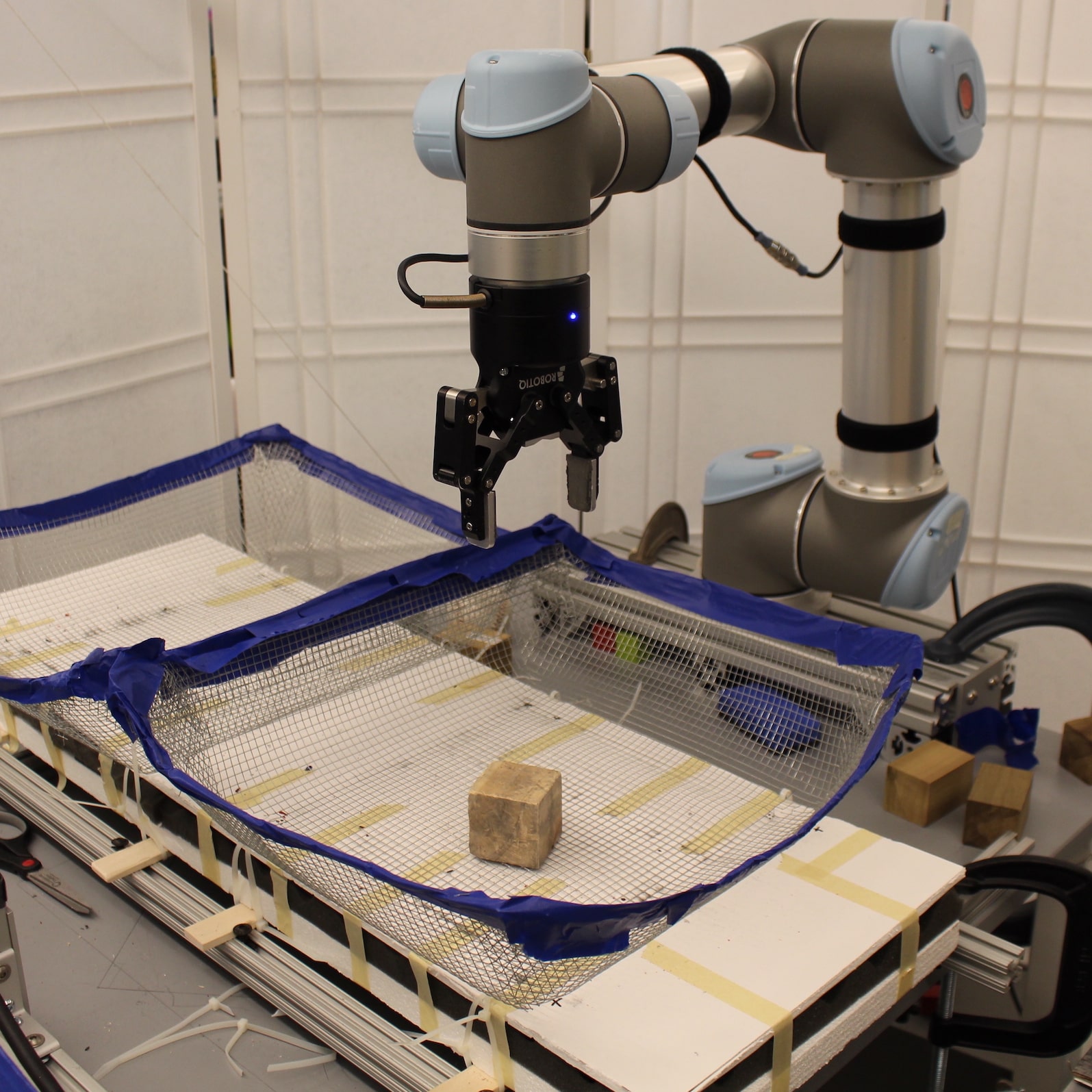}
\includegraphics[width=\env]{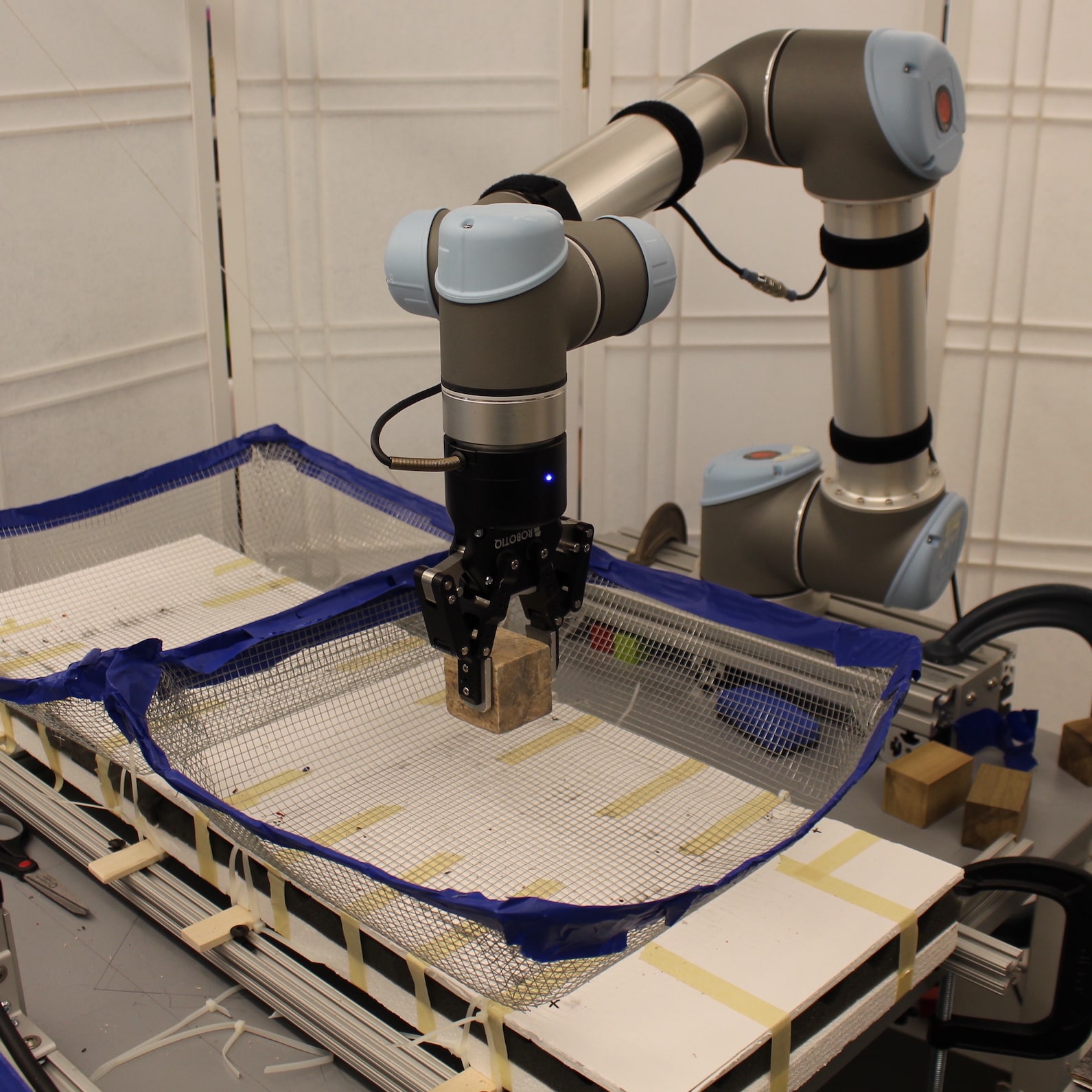}
}
\subfloat[Clutter Grasping]{
\includegraphics[width=\env]{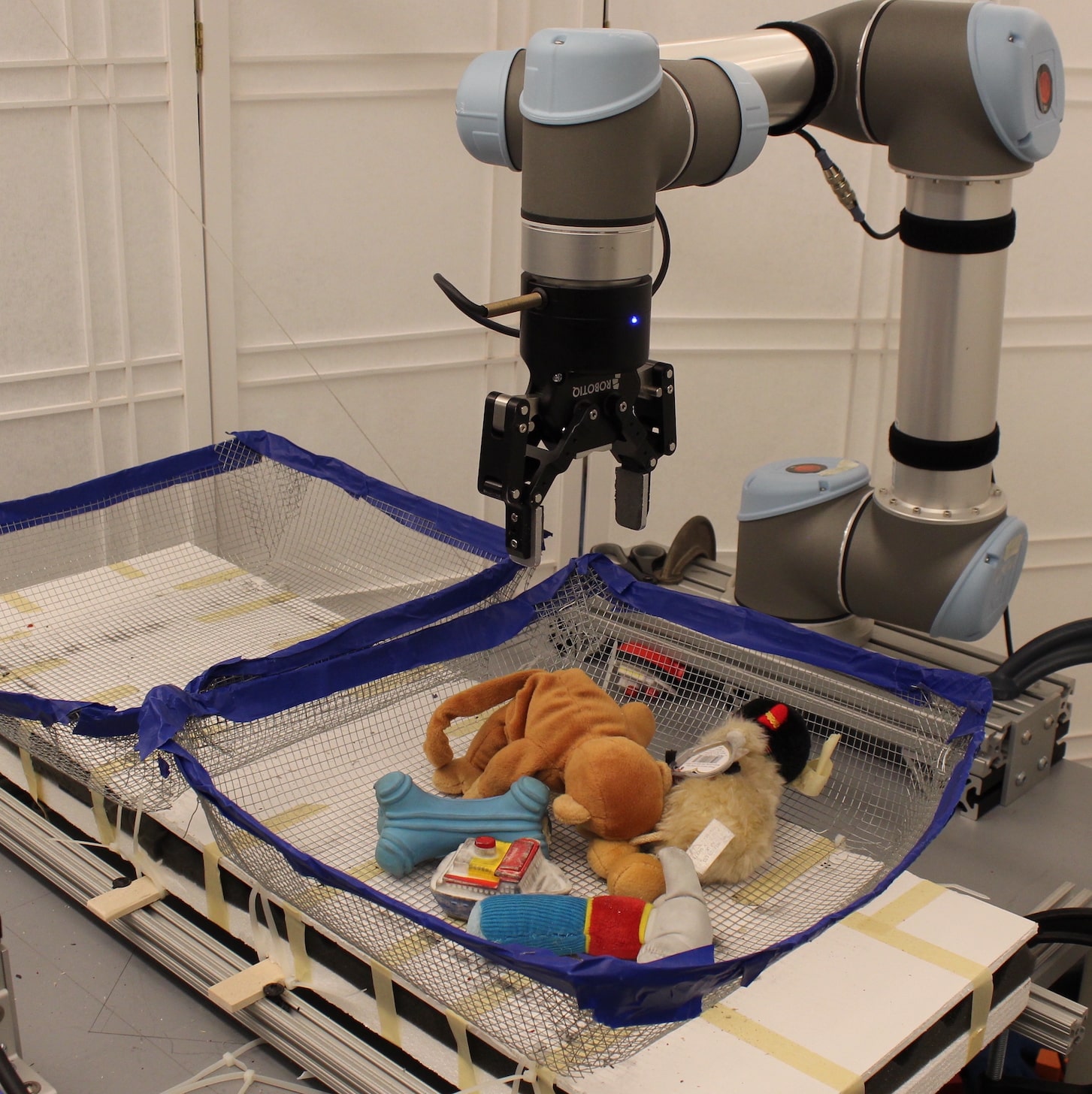}
\includegraphics[width=\env]{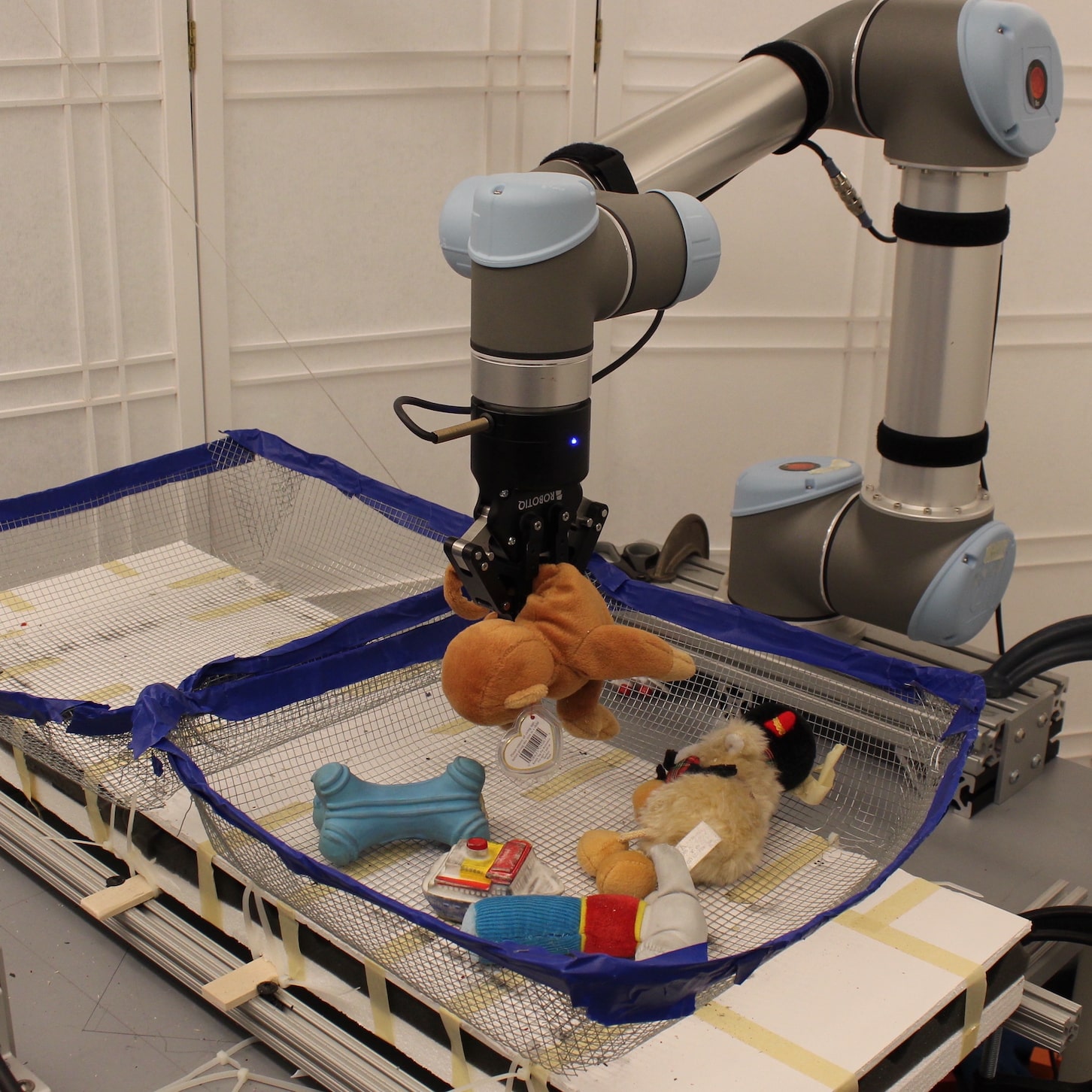}
}
\subfloat[Block Pushing]{
\includegraphics[width=\env]{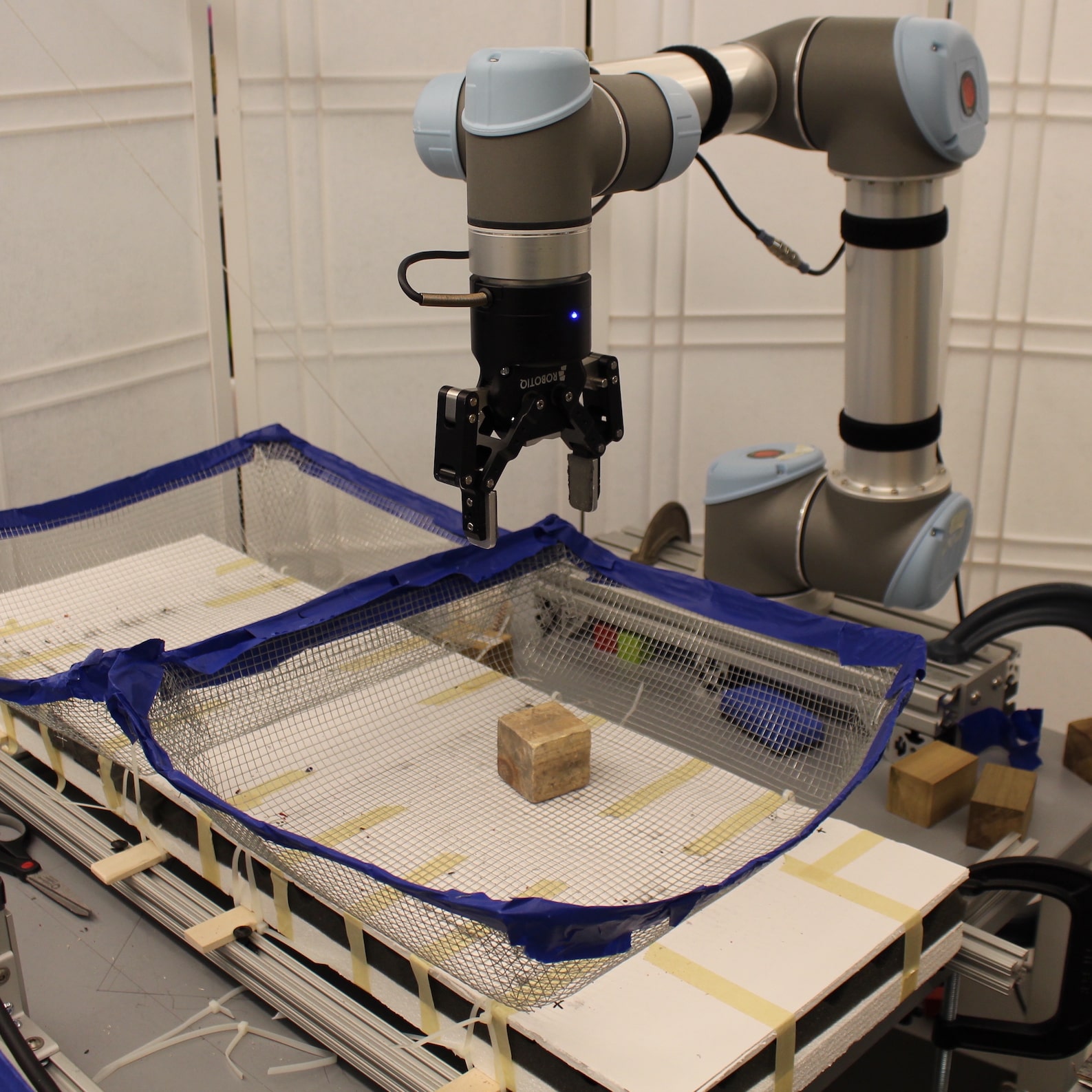}
\includegraphics[width=\env]{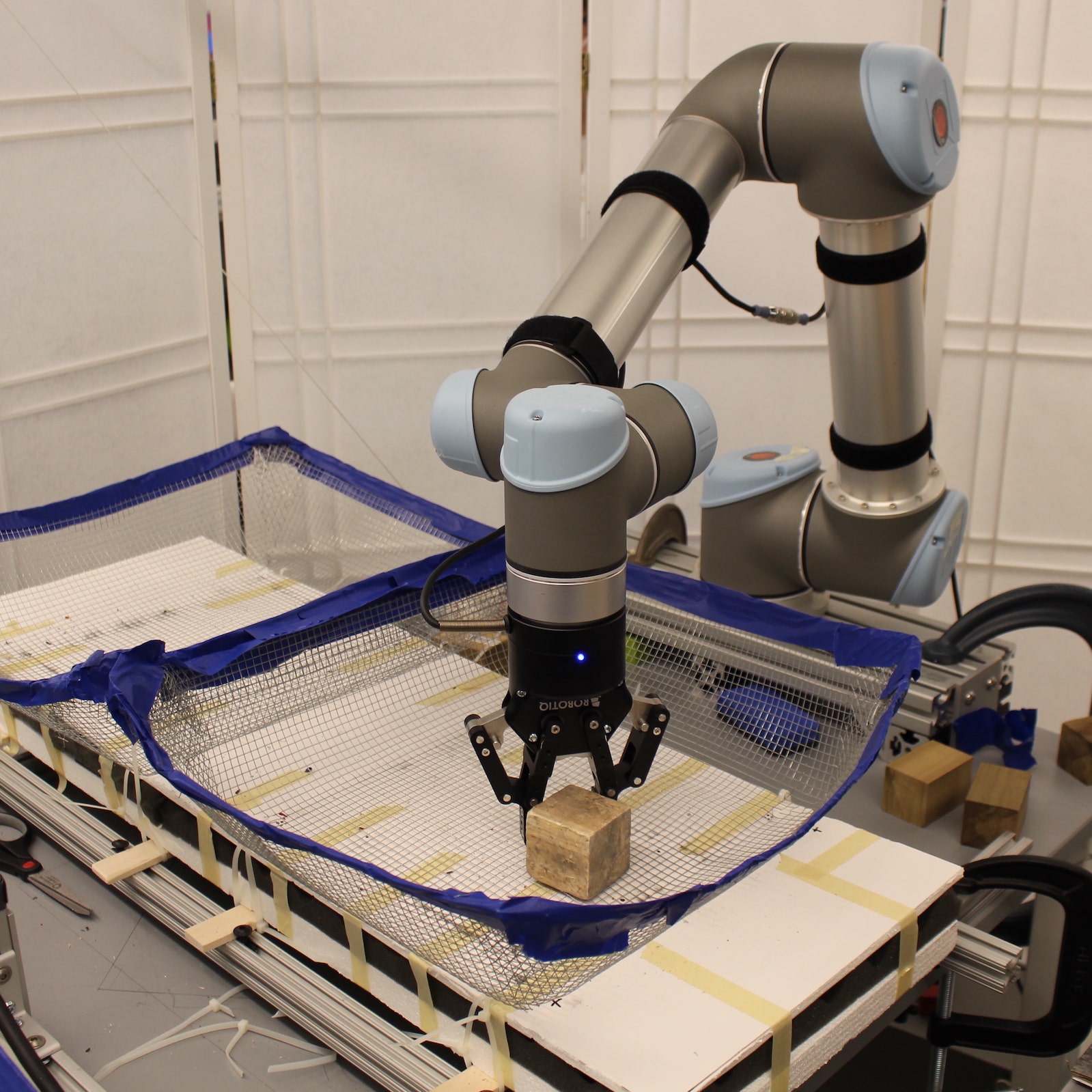}
}
\subfloat[Block in Bowl]{
\includegraphics[width=\env]{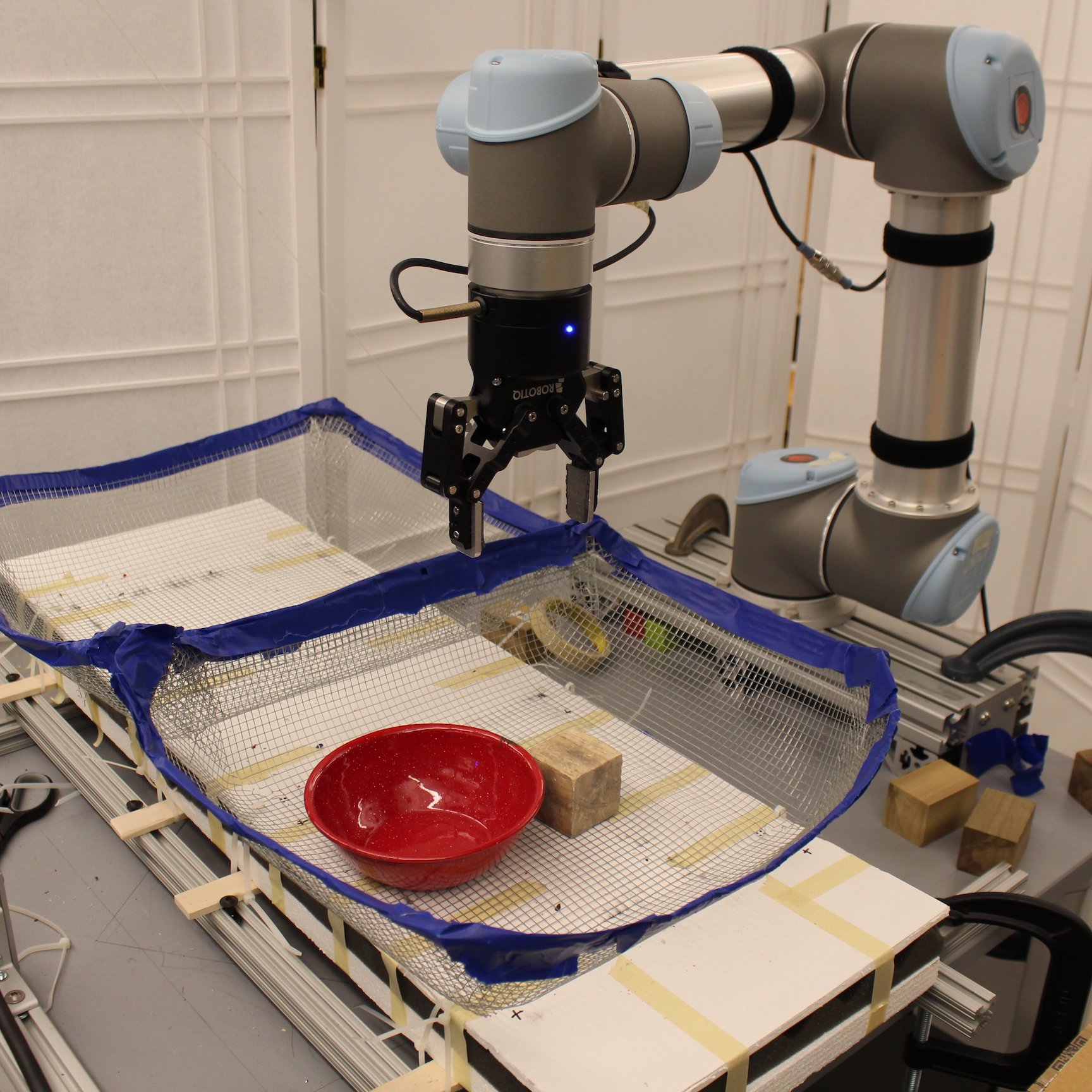}
\includegraphics[width=\env]{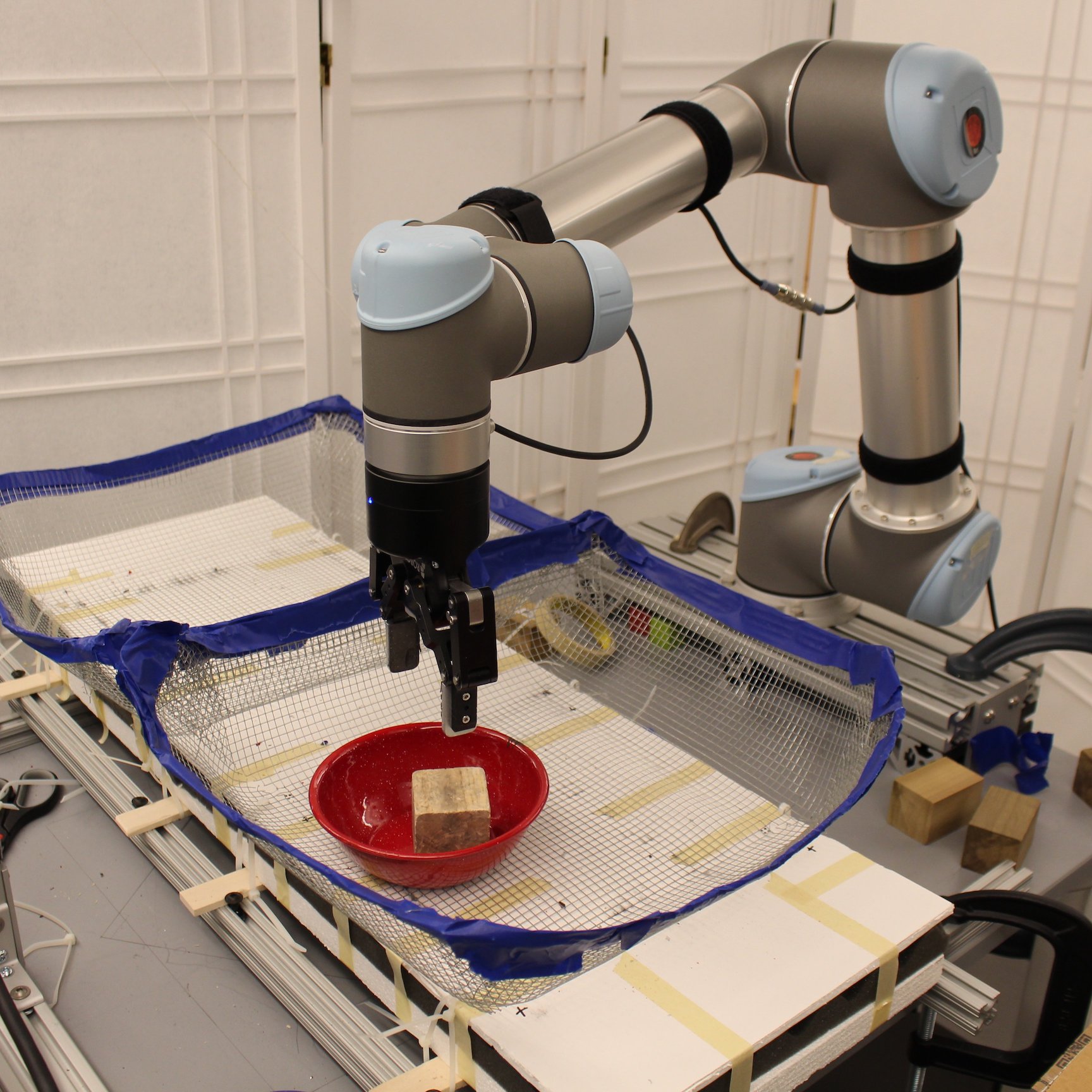}}
\caption{(a)-(d): Our simulation environments implemented in PyBullet~\cite{pybullet}. The left images in each environment show the initial state of the environment; the right images in each environment show the goal state. (e)-(h): Our on-robot learning environments.}
\label{fig:env}
\end{figure*}

The specific choice of symmetry group and data augmentation strategy have a significant impact on the sample efficiency of the algorithm. Before evaluating our algorithms on the robot, we evaluate those different algorithmic choices in simulation. Here, we experiment in the context of Equivariant SAC for the four tabletop manipulation tasks shown in Figure~\ref{fig:env}. 

All tasks have sparse rewards, i.e., +1 reward for reaching the goal, and 0 otherwise. We use a 2-channel image as the observation. The first channel is a top-down depth image centered with respect to the robot gripper. The gripper is drawn at the center of the depth image with its current aperture and orientation. The second channel is a binary channel (i.e., the values of all pixels are either 0 or 1), indicating if the gripper is holding an object. The action space is: $x, y, z \in [-0.05m, 0.05m]; \theta \in [-\frac{\pi}{4}, \frac{\pi}{4}]; \lambda \in [0, 1]$ (0 means fully close and 1 means fully open). 20 episodes of expert demonstration are added to the replay buffer before the start of training (see the ablation study about the effect of expert demonstrations in Appendix~\ref{appendix:expert}). 
See the detailed description of the environments in Appendix~\ref{app:sim_env} and the training details in Appendix~\ref{app:traininig_detail}.



\begin{figure}[t]
\centering
\subfloat[Comparison of Equivariant SAC defined with different symmetry groups.]{
\includegraphics[width=0.98\textwidth]{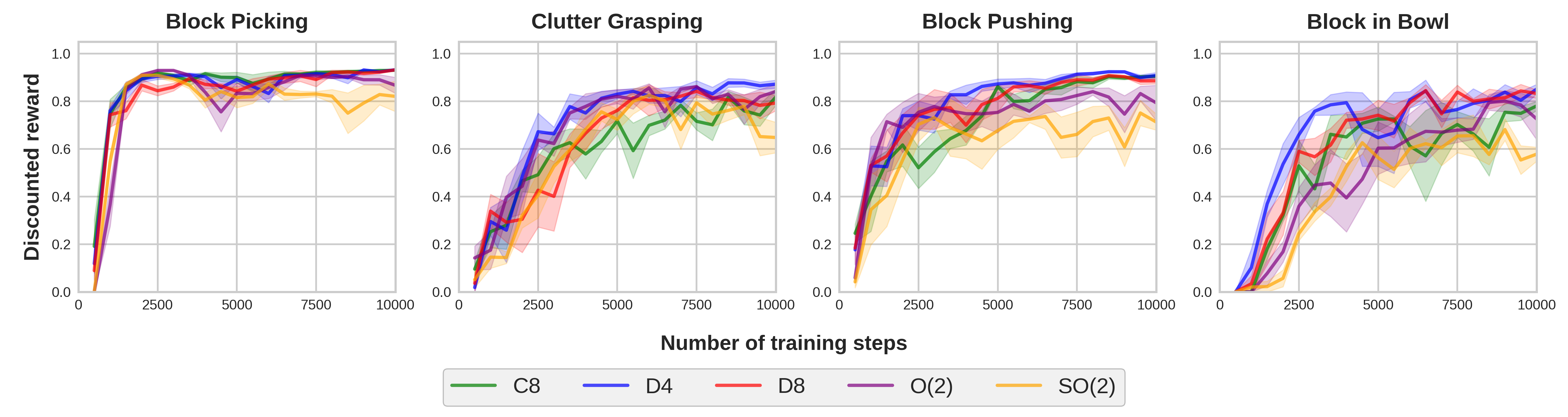}
\label{fig:exp_group}
}\\
\subfloat[Comparison of Equivariant SAC equipped with different data augmentation techniques.]{
\includegraphics[width=0.98\textwidth]{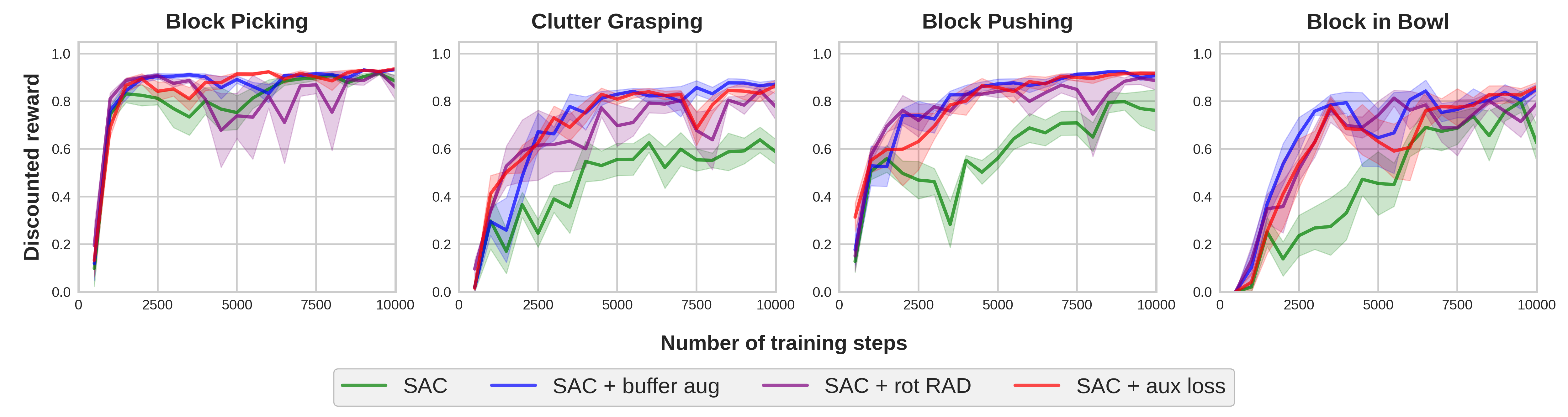}
\label{fig:exp_aug}
}
\caption{\plotDescription}
\end{figure}

\subsection{Choice of Symmetry Group}
\label{sec:exp_group}


In Equivariant SAC, we must select a symmetry group with which to parameterize the actor and critic models. This symmetry group reflects the underlying symmetry that we assume to exist in the problem. One might expect that larger symmetry groups would always be better because they would enable the model to generalize to a greater number of different situations. However, as shown in \citet{e2cnn}, a larger group does not necessarily improve the performance of the model. It is therefore important to ask which symmetry groups are most helpful in our robotic manipulation domains. We compare the performance of Equivariant SAC when the actor/critic models are parameterized by each of the following five different symmetry groups: 1) $C_8$: the cyclic group that encodes discrete rotations every 45 degrees; 2) $D_4$: the Dihedral group that encodes rotations every 90 degrees and reflection; 3) $D_8$: the Dihedral group that encodes rotations every 45 degrees and reflection; 4) $\SO(2)$: the group of continuous planar rotations; 5) $\mathrm{O}(2)$: the group of continuous planar rotations and reflections. \edit{In all baselines, we use data augmentation in the replay buffer where four extra transitions with random $\SO(2)$ is generated for each experienced transition.}


Figure~\ref{fig:exp_group} shows the result. 
\edit{In all four environments, $D_4$ and $D_8$ show the best performance over the five different groups in terms of convergence speed and converged performance, where $D_4$ has a marginal improvement over $D_8$ even though $D_8$ encodes more rotations than $D_4$.}
In addition, we find that incorporating reflection symmetry generally improves the performance (comparing $D_4$ vs $C_8$ and $\mathrm{O}(2)$ vs $\SO(2)$). However, both the $\SO(2)$ and $\mathrm{O}(2)$ networks underperform the $D_4$ network. We hypothesize that this is because the $D_4$ network has access to the regular representation of the symmetry group as the hidden layer of the network, whereas the models defined over the continuous groups $\SO(2)$ and $\OO(2)$ do not. The regular representation explicitly encodes the feature maps over all elements in the finite group concurrently, making it an informative representation for the hidden layers. Moreover, the regular representation is compatible with component-wise activation functions and component-wise max pooling, which are most commonly used and known to work well in deep networks.


\subsection{Choice of Data Augmentation Strategy}
\label{sec:exp_aug}
The results of Section~\ref{sec:exp_group} indicate that the equivariant models that perform best in our domains use the $D_4 \subset \mathrm{O}(2)$ symmetry group, the group of $90$ degree rotations and reflections. As a result, it could still make sense to use data augmentation to enable the model to learn rotational symmetry within a continuous range $(0,\frac{\pi}{2})$. Here, we compare three different approaches to data augmentation with random rotations: SAC + buffer aug, SAC + aux loss, and SAC + rot RAD. In \emph{SAC + buffer aug}, we add four extra transitions with random $\SO(2)$ rotations to the replay buffer for each experienced transition. (See Appendix~\ref{appendix:exp_buffer_aug} for an comparison of different amounts of augmentation.) In \emph{SAC + aux loss}, we add extra loss terms to encourage the model to learn continuous rotational equivariant representations:  
$\mathcal{L}_\mathrm{aux}^\mathrm{actor} = \frac{1}{2}(\pi(g\mathcal{F}_s) - g\pi(\mathcal{F}_s))^2$ and  $\mathcal{L}_\mathrm{aux}^\mathrm{critic} = \frac{1}{2}(q(g\mathcal{F}_s, ga) - q(\mathcal{F}_s, a))^2$, 
where $g\in \SO(2)$ and $(\mathcal{F}_s, a)$ is the sampled state-action pair in the minibatch. In \emph{SAC + rot RAD}, we use RAD~\cite{rad} to perform rotational data augmentation at each sample and training step. Figure~\ref{fig:exp_aug} shows the comparison between the three data augmentation approaches and vanilla Equivariant SAC (green). Even though Equivariant SAC already encodes D4 equivariance within the structure of the network, adding $\SO(2)$ data augmentations to the algorithm can still help by a substantial margin, especially in the more challenging tasks like clutter grasping, block pushing, and block in bowl. However, notice that all three data augmentations approaches perform similarly. As such, we adopt the buffer augmentation method in the remainder of this paper because it is the simplest method.

\section{On-Robot Learning}




\begin{wrapfigure}[19]{r}{0.7\textwidth}
\vspace{-0.3cm}
\centering
\includegraphics[width=\linewidth]{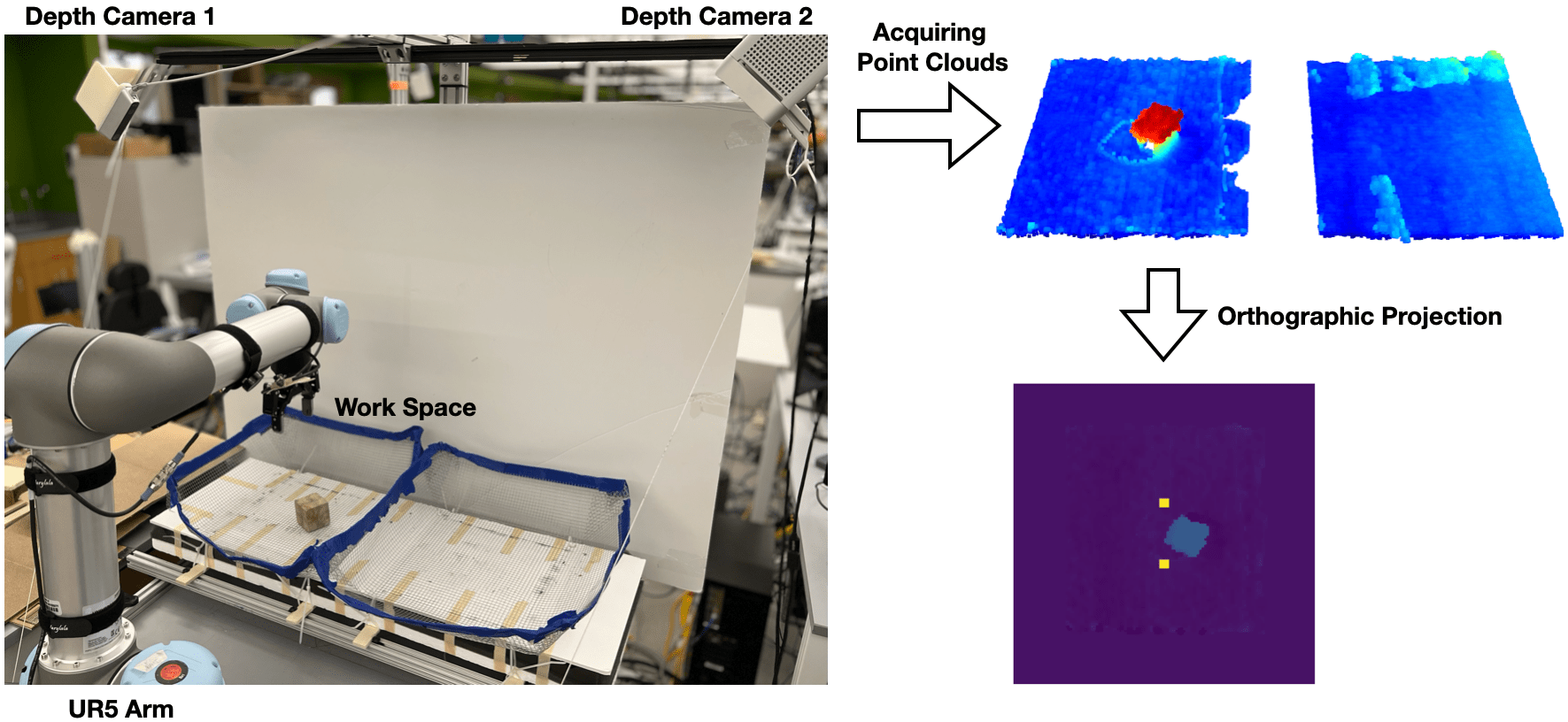}
\vspace{-0.5cm}
\caption{Our experimental set up for on-robot learning. The observation (bottom right) is generated by first acquiring point clouds from two depth cameras above the workspace then creating an orthographic projection at the gripper's position. The gripper is drawn at the center of the observation (in yellow) with its current aperture and orientation.}
\label{fig:workspace}
\end{wrapfigure}

Figure~\ref{fig:workspace} illustrates our on-robot learning setup. We use a Universal Robots UR5 arm equipped with a Robotiq 2F-85 parallel-jaw gripper. Since we need to be able to see the objects on the table beneath the robot hand and arm, we mount two depth sensors that view the scene from orthogonal directions each at 45 degrees with respect to the bins (see Figure~\ref{fig:workspace} left). One of these sensors is an Occipital Structure Sensor and the other is a Microsoft Azure Kinect DK. The output of each sensor is converted to a partial point cloud, fused into a single combined point cloud, and projected into a depth image viewed from a top-down direction (Figure~\ref{fig:workspace} right). \edit{We use the same 2-channel observation as in Section~\ref{sec:sim_exp} including a depth channel and a binary channel for the gripper state.} Our workstation has a Intel Core i7-9700k CPU (3.60GHz) and a Nvidia RTX 2080Ti GPU. All together, this setup enables us to train at a rate of approximately 1.1 seconds per environmental transition. This includes the time it takes to take a single transition on the physical robot as well as to perform a single SGD step in the model (we do this in parallel with the robot motion), but it does not include the time it takes to reset the environment between episodes.

\subsection{Physical Tasks}

All physical experiments are performed using the same four tasks 
described in Section~\ref{sec:sim_exp} (see Figure~\ref{fig:env})~\footnote{Note that in Block Pushing, we use a virtual goal drawn on the observation, so the goal is visible to the agent but not to the human observer.}. In tasks involving only a single object (i.e., Block Picking and Block Pushing), the robot uses only one bin as the workspace. In tasks involving multiple objects (i.e., Clutter Grasping and Block in Bowl), the robot alternately uses one of the two bins as the active workspace and then uses the other bin to reset the environment. We implement an automated resetting process for all four environments. In Block Picking and Block Pushing, the robot will reset the environment through picking up the block and randomly placing it inside the workspace. In Block in Bowl, the robot will move both the bowl and the block from the active workspace to random positions in the reset bin, then switch the active workspace to the reset bin. In Clutter Grasping, after the robot successfully grasps one object from the active workspace, it will drop the object to a random position in the reset bin. Once the robot grasps all objects from the active workspace, the robot will switch the active workspace to the reset bin. See Appendix~\ref{app:real_env} for more details.



\begin{table*}[t]
\small
    \centering
    \begin{tabular}{ccccc}
    \toprule
    Task & Block Picking & Clutter Grasping & Block Pushing & Block in Bowl \\
    \midrule
    \midrule
    Number of training steps & 2000 & 2000 & 2000 & 4000\\
    \midrule
    Approximate time for training & 45 mins & 45 mins & 1 hr & 2 hrs 40 mins\\
    \midrule 
    Evaluation success rate & 100\% (50/50) & 96\% (48/50) & 92\% (46/50) & 92\% (46/50)\\
    \midrule
    \midrule
    Sim2real transfer success rate & 100\%(50/50) & 86\%(43/50) & 70\%(35/50) & 72\%(36/50) \\
    \bottomrule
    \end{tabular}
    \caption{Top: the number of training steps, approximate time for training, and the evaluation success rate of the trained policy of our on-robot learning. Bottom: the evaluation success rate of the model trained in simulation. All succes rates are averaged over 50 episodes.}
    \label{tab:on_robot_result}
\end{table*}


\subsection{Performance of the On-Robot Learned Policy}

First, we evaluate the ability of Equivariant SAC to learn these tasks entirely in the on-robot setting, i.e., entirely on the robot without pre-training. Based on the findings described in Section~\ref{sec:sim_exp}, we configure both our actor and critic models to use the $D_4$ symmetry group, and we use the ``SAC + buffer aug'' strategy described in Section~\ref{sec:exp_aug}. The blue line in Figure~\ref{fig:exp_real} shows the learning curve of training, and the third row of Table~\ref{tab:on_robot_result} shows the greedy performance of the learned policy after training, averaged over 50 test episodes.

In Block Picking, the robot succeeded in all 50 trials. In Clutter Grasping, the robot failed to find an appropriate grasp point in two out of the 50 episodes. In Block Pushing, the robot failed in four out of 50 episodes. In two of these, it failed to move towards the block. In the other two failures, the robot kept pushing down from the top of the block and triggered the UR5 safety system. In Block in Bowl, the robot failed in four out of the 50 trials. In three of these, the robot grasped the block but did not move towards the bowl. In the other failure, the robot grasped the bowl and the block together in a single grasp and failed to let go.

In Block Picking, Clutter Grasping, and Block Pushing, training lasts 2000 transitions. In Block in Bowl, training lasts 4000 transitions because it is a harder task to learn. In terms of wall clock time, training tasks a total of 45 minutes in Block Picking and Clutter Grasping; one hour in Block Pushing (because of the additional time required to physically reset that environment); and 2 hours and 40 minutes in Block in Bowl. For Block in Bowl, this time can be decomposed into approximately 73 minutes ($1.1s \times 4000$) of environmental steps and 107 minutes to reset the environment.




\subsection{Baseline Comparison}

There are very few methods in the literature that can learn efficiently in the on-robot setting. Here, we baseline against the Framework for Efficient Robotic Manipulation (FERM)~\cite{ferm}. FERM utilizes contrastive learning with random crop augmentations to improve sample efficiency. We implement FERM such that the image encoder has a similar amount of trainable parameters as the equivariant network (1.5M vs 1.1M). See Appendix~\ref{app:baseline_detail} for details. Figure~\ref{fig:exp_real} compares the learning curves of Equivariant SAC (blue) and FERM (red) where both methods learn on-robot. In Block Picking, Equivariant SAC masters the task after about 1000 steps, while FERM learns much more slowly. In Clutter Grasping, FERM performs better at the beginning phase of learning but fails to converge to a good policy at the end of learning. We hypothesize that this is because: 1) the pre-training of the encoder in FERM helps the network to learn a better feature representation at the early phase of learning. 2) this task does not require accurate manipulation as the other three tasks since the objects are all deformable (this also explains why the on-robot learning is faster than the simulation in this task). In Block Pushing and Block in Bowl, FERM fails to learn any good policy at all, while Equivariant SAC solves the task within 2000 and 4000 steps, respectively. 



\begin{figure}[t]
\centering
\includegraphics[width=\textwidth]{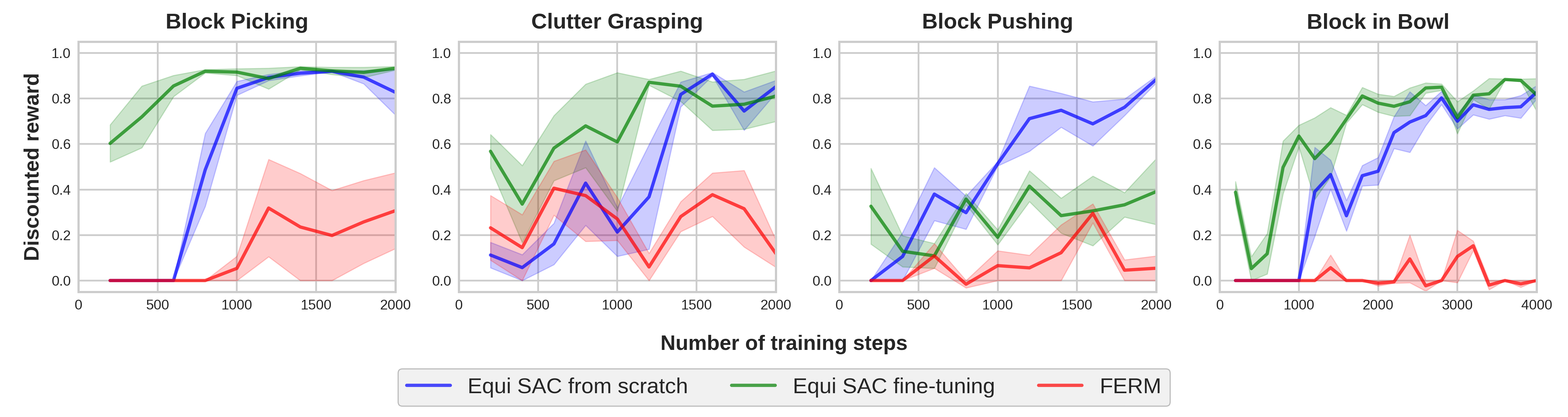}
\caption{Comparison of Equivariant SAC trained from scratch (blue), Equivariant SAC with sim-to-real fine-tuning (green), and FERM (red) in real world. The plots show the performance of the behavior policy in terms of the discounted reward. Each point is the average discounted reward in the previous 200 steps. Results are averaged over three runs. Shading denotes standard error.}
\label{fig:exp_real}
\end{figure}

\begin{figure}
\centering
\subfloat[]{\includegraphics[height=0.18\textwidth]{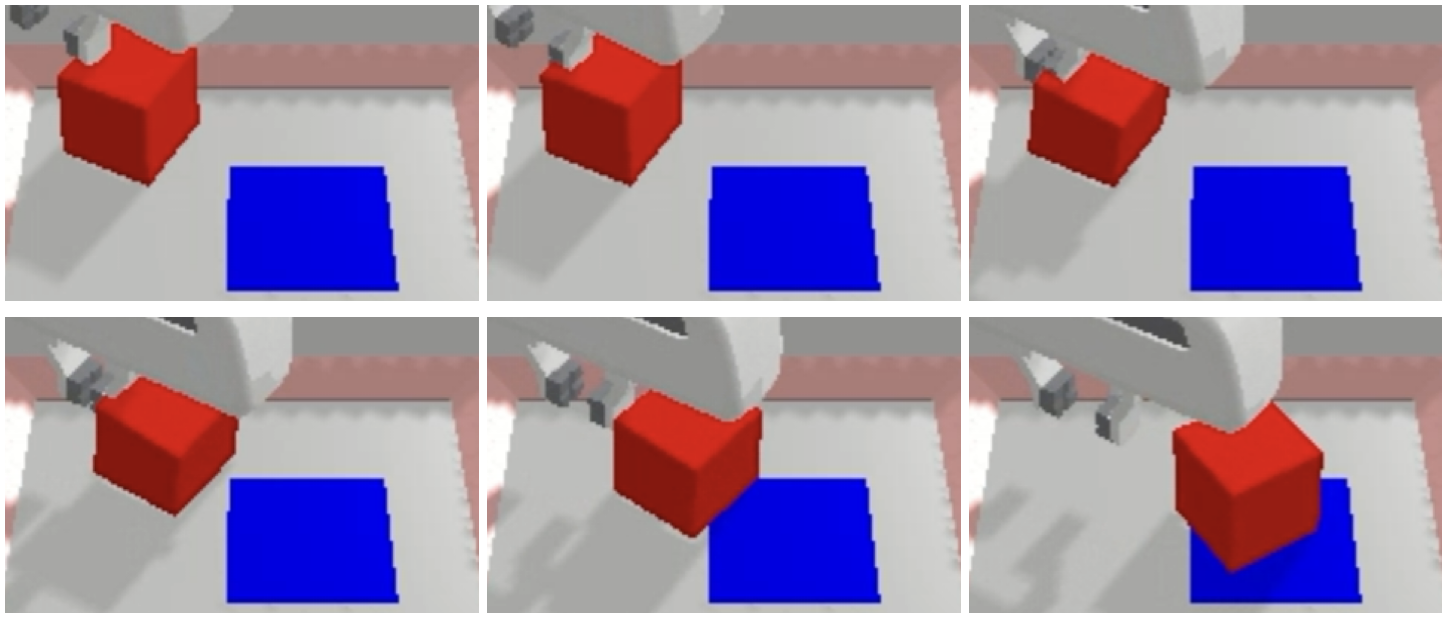}
}
\subfloat[]{
\includegraphics[width=0.18\textwidth]{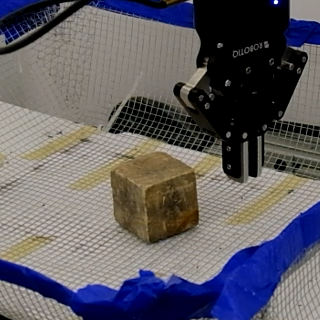}
\includegraphics[width=0.18\textwidth]{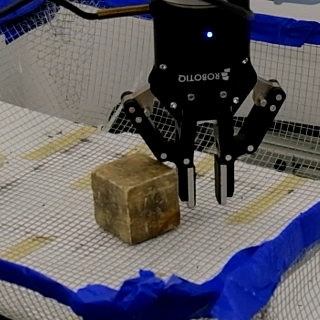}
\includegraphics[width=0.18\textwidth]{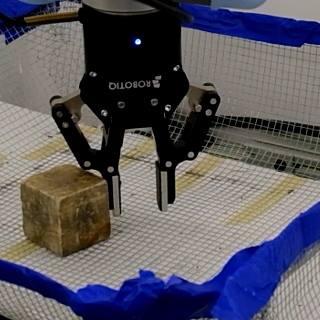}
}
\caption{The different strategy between the simulation agent and the on-robot learning agent. (a): In simulation, the agent presses on the edge of the block and bounces it away. (b): In the real world, the agent pushes the block.}
\label{fig:push_sim_vs_real}
\end{figure}

\subsection{Sim2real Comparison}

Here, we compare direct on-robot learning with a sim2real strategy where we first train the agent to convergence in simulation and then transfer to the physical system by directly copying model parameters. We used the PyBullet simulation environment described in Section~\ref{sec:sim_exp}. 
The last row of Table~\ref{tab:on_robot_result} shows the performance of the agent trained in simulation when that policy is executed on the physical robot. In general, the sim2real policy  performs significantly worse than the on-robot trained policy: 86\% versus 96\% in Clutter Grasping, 70\% versus 92\% in Block Pushing, and 72\% versus 92\% in Block in Bowl.  The exception is Block Picking in which both the sim2real and on-robot policies achieve 100\% success.
This result explicitly reflects the sim2real performance gap.





\subsection{Fine-tuning Sim2real on Physical Robots}

We also evaluated our ability to improve the policy learned in simulation by fine-tuning on the physical robot. The green line in Figure~\ref{fig:exp_real} shows the learning curve of this fine-tuning agent. Before fine-tuning begins, the models used by this agent are loaded with the model parameters learned in simulation. This should be compared with the performance of the on-robot agent (no pre-training), shown in blue in Figure~\ref{fig:exp_real}. In Block Picking, Clutter Grasping and Block in Bowl, both the training from scratch and sim2real fine-tuning converge to the same level of performance at the end of training, where the training from scratch agent is about 500 steps slower than the fine-tuning agent in terms of converging speed. This suggests that while pre-training helps in these tasks, it is not critical to successfully learning the policy. 

In Block Pushing, on the other hand, the fine-tuning agent performs very poorly and it does not recover during training (green versus blue line in Figure~\ref{fig:exp_real}). To understand why this occurs, we checked the policies typically learned by the different agents for Block Pushing. Whereas the simulation agent learns to move the block by pushing down on the block \emph{from the top} (see Figure~\ref{fig:push_sim_vs_real} (a)), the on-robot agent learns a policy that pushes the block \emph{from the side} (see Figure~\ref{fig:push_sim_vs_real} (b)). We hypothesize that this is what impedes the sim2real fine-tuning agent on this task. In order to succeed on the physical robot, the sim2real agent must first \emph{unlearn} the policy that was successful in simulation. This is an example of negative transfer where the pre-trained policy actually impedes on-robot learning.
The fact that the simulation agent learns to push down on the block (and then to push sideways) whereas the on-robot agent learns to push from the side is probably the result of differences between the contact dynamics of the simulator and the real robot. The PyBullet simulator has slightly softer contact compliance that enables it to push down without generating too much force and triggering the UR5 safety system. Also, the friction coefficient between the gripper and the block is larger than between the block and the ground, allowing the simulator to slide the block. Notice that it would be very hard to close this gap by improving the simulator because it is difficult to measure and model physical friction and compliance accurately.

\section{Discussion}


This paper proposes an approach to on-robot learning using Equivariant SAC in combination with data augmentation that can learn to solve simple manipulation tasks in a couple of hours without pre-training. Using the new method, we find that it may sometimes be unnecessary to pre-train in simulation before training on a physical robot and that pre-training in simulation can sometimes be harmful because it causes negative transfer.

\section{Limitations}

A key limitation of the approach is that the use of equivariant models requires making assumptions about which symmetries are present in the domain. While many robotics problems have rotation, translation, and reflection symmetries, these symmetries may not be present in all regions of the state space. The current approach requires the system designer to model these asymmetries explicitly. However, the system would ideally have the ability to learn where symmetries are present without hand coding. Similarly, many problems have domain-specific symmetries beyond rotation and reflection symmetries that can be hard for the system designer to recognize. We would like our system to be able to identify which symmetries are present and then incorporate them into the relevant neural models. Another area for future work is to understand when incorrect symmetry assumptions are still useful. We have observed that equivariant models can still speed up learning, even when the symmetry assumptions are sometimes violated in a specific domain. We would like to have a theory for understanding this phenomenon more precisely. \edit{Third, though our method is not limited to depth images, this work demonstrates equivariant learning using depth images only because our tasks do not require RGB information. In the future work, we will incorporate RGB image to improve the robustness of our method to transparent or reflective objects.}

\section*{Acknowledgments}
This work is supported in part by NSF 1724257, NSF 1724191, NSF 1763878, NSF 1750649, and NASA 80NSSC19K1474. R. Walters is supported by the Roux Institute and the Harold Alfond Foundation and NSF grants 2107256 and 2134178.


\bibliography{main}

\begin{thebibliography}{36}
\providecommand{\natexlab}[1]{#1}
\providecommand{\url}[1]{\texttt{#1}}
\expandafter\ifx\csname urlstyle\endcsname\relax
  \providecommand{\doi}[1]{doi: #1}\else
  \providecommand{\doi}{doi: \begingroup \urlstyle{rm}\Url}\fi

\bibitem[Pinto and Gupta(2016)]{pinto2016supersizing}
L.~Pinto and A.~Gupta.
\newblock Supersizing self-supervision: Learning to grasp from 50k tries and
  700 robot hours.
\newblock In \emph{2016 IEEE international conference on robotics and
  automation (ICRA)}, pages 3406--3413. IEEE, 2016.

\bibitem[Kalashnikov et~al.(2018)Kalashnikov, Irpan, Pastor, Ibarz, Herzog,
  Jang, Quillen, Holly, Kalakrishnan, Vanhoucke, et~al.]{qt_opt}
D.~Kalashnikov, A.~Irpan, P.~Pastor, J.~Ibarz, A.~Herzog, E.~Jang, D.~Quillen,
  E.~Holly, M.~Kalakrishnan, V.~Vanhoucke, et~al.
\newblock Qt-opt: Scalable deep reinforcement learning for vision-based robotic
  manipulation.
\newblock \emph{arXiv preprint arXiv:1806.10293}, 2018.

\bibitem[Berscheid et~al.(2021)Berscheid, Friedrich, and
  Kr{\"o}ger]{berscheid2021robot}
L.~Berscheid, C.~Friedrich, and T.~Kr{\"o}ger.
\newblock Robot learning of 6 dof grasping using model-based adaptive
  primitives.
\newblock In \emph{2021 IEEE International Conference on Robotics and
  Automation (ICRA)}, pages 4474--4480. IEEE, 2021.

\bibitem[van~der Pol et~al.(2020)van~der Pol, Worrall, van Hoof, Oliehoek, and
  Welling]{van2020mdp}
E.~van~der Pol, D.~Worrall, H.~van Hoof, F.~Oliehoek, and M.~Welling.
\newblock Mdp homomorphic networks: Group symmetries in reinforcement learning.
\newblock \emph{Advances in Neural Information Processing Systems}, 33, 2020.

\bibitem[Wang et~al.(2022)Wang, Walters, and Platt]{iclr}
D.~Wang, R.~Walters, and R.~Platt.
\newblock {$\mathrm{SO}(2)$}-equivariant reinforcement learning.
\newblock In \emph{International Conference on Learning Representations}, 2022.
\newblock URL \url{https://openreview.net/forum?id=7F9cOhdvfk_}.

\bibitem[H{\"o}fer et~al.(2020)H{\"o}fer, Bekris, Handa, Gamboa, Golemo,
  Mozifian, Atkeson, Fox, Goldberg, Leonard, et~al.]{hofer2020perspectives}
S.~H{\"o}fer, K.~Bekris, A.~Handa, J.~C. Gamboa, F.~Golemo, M.~Mozifian,
  C.~Atkeson, D.~Fox, K.~Goldberg, J.~Leonard, et~al.
\newblock Perspectives on sim2real transfer for robotics: A summary of the r:
  Ss 2020 workshop.
\newblock \emph{arXiv preprint arXiv:2012.03806}, 2020.

\bibitem[Weiler and Cesa(2019)]{e2cnn}
M.~Weiler and G.~Cesa.
\newblock General $ e (2) $-equivariant steerable cnns.
\newblock \emph{arXiv preprint arXiv:1911.08251}, 2019.

\bibitem[Cohen and Welling(2016{\natexlab{a}})]{g_conv}
T.~Cohen and M.~Welling.
\newblock Group equivariant convolutional networks.
\newblock In \emph{International conference on machine learning}, pages
  2990--2999. PMLR, 2016{\natexlab{a}}.

\bibitem[Cohen and Welling(2016{\natexlab{b}})]{steerable_cnns}
T.~S. Cohen and M.~Welling.
\newblock Steerable cnns.
\newblock \emph{arXiv preprint arXiv:1612.08498}, 2016{\natexlab{b}}.

\bibitem[Benton et~al.(2020)Benton, Finzi, Izmailov, and
  Wilson]{benton2020learning}
G.~Benton, M.~Finzi, P.~Izmailov, and A.~G. Wilson.
\newblock Learning invariances in neural networks.
\newblock \emph{arXiv preprint arXiv:2010.11882}, 2020.

\bibitem[Dey et~al.(2020)Dey, Chen, and Ghafurian]{dey2020group}
N.~Dey, A.~Chen, and S.~Ghafurian.
\newblock Group equivariant generative adversarial networks.
\newblock \emph{arXiv preprint arXiv:2005.01683}, 2020.

\bibitem[Wang et~al.(2020)Wang, Walters, and Yu]{wang2020incorporating}
R.~Wang, R.~Walters, and R.~Yu.
\newblock Incorporating symmetry into deep dynamics models for improved
  generalization.
\newblock \emph{arXiv preprint arXiv:2002.03061}, 2020.

\bibitem[Walters et~al.(2020)Walters, Li, and Yu]{walters2020trajectory}
R.~Walters, J.~Li, and R.~Yu.
\newblock Trajectory prediction using equivariant continuous convolution.
\newblock \emph{arXiv preprint arXiv:2010.11344}, 2020.

\bibitem[Mondal et~al.(2020)Mondal, Nair, and Siddiqi]{mondal2020group}
A.~K. Mondal, P.~Nair, and K.~Siddiqi.
\newblock Group equivariant deep reinforcement learning.
\newblock \emph{arXiv preprint arXiv:2007.03437}, 2020.

\bibitem[Wang et~al.(2021)Wang, Walters, Zhu, and Platt]{corl}
D.~Wang, R.~Walters, X.~Zhu, and R.~Platt.
\newblock Equivariant \$q\$ learning in spatial action spaces.
\newblock In \emph{5th Annual Conference on Robot Learning}, 2021.
\newblock URL \url{https://openreview.net/forum?id=IScz42A3iCI}.

\bibitem[Zhu et~al.(2022)Zhu, Wang, Biza, Su, Walters, and Platt]{zhu2022grasp}
X.~Zhu, D.~Wang, O.~Biza, G.~Su, R.~Walters, and R.~Platt.
\newblock Sample efficient grasp learning using equivariant models.
\newblock \emph{Proceedings of Robotics: Science and Systems (RSS)}, 2022.

\bibitem[Huang et~al.(2022)Huang, Wang, Walters, and Platt]{equi_transporter}
H.~Huang, D.~Wang, R.~Walters, and R.~Platt.
\newblock Equivariant transporter network.
\newblock In \emph{Proceedings of Robotics: Science and Systems (RSS)}, 2022.

\bibitem[Rusu et~al.(2017)Rusu, Ve{\v{c}}er{\'\i}k, Roth{\"o}rl, Heess,
  Pascanu, and Hadsell]{rusu2017sim}
A.~A. Rusu, M.~Ve{\v{c}}er{\'\i}k, T.~Roth{\"o}rl, N.~Heess, R.~Pascanu, and
  R.~Hadsell.
\newblock Sim-to-real robot learning from pixels with progressive nets.
\newblock In \emph{Conference on Robot Learning}, pages 262--270. PMLR, 2017.

\bibitem[Zhu et~al.(2018)Zhu, Wang, Merel, Rusu, Erez, Cabi, Tunyasuvunakool,
  Kram{\'a}r, Hadsell, de~Freitas, et~al.]{zhu2018reinforcement}
Y.~Zhu, Z.~Wang, J.~Merel, A.~Rusu, T.~Erez, S.~Cabi, S.~Tunyasuvunakool,
  J.~Kram{\'a}r, R.~Hadsell, N.~de~Freitas, et~al.
\newblock Reinforcement and imitation learning for diverse visuomotor skills.
\newblock \emph{arXiv preprint arXiv:1802.09564}, 2018.

\bibitem[Gualtieri and Platt(2020)]{gualtieri2020learning}
M.~Gualtieri and R.~Platt.
\newblock Learning manipulation skills via hierarchical spatial attention.
\newblock \emph{IEEE Transactions on Robotics}, 36\penalty0 (4):\penalty0
  1067--1078, 2020.

\bibitem[Wang et~al.(2020)Wang, Kohler, and Platt]{wang2020policy}
D.~Wang, C.~Kohler, and R.~Platt.
\newblock Policy learning in se (3) action spaces.
\newblock \emph{arXiv preprint arXiv:2010.02798}, 2020.

\bibitem[Biza et~al.(2021)Biza, Wang, Platt, van~de Meent, and
  Wong]{biza2021action}
O.~Biza, D.~Wang, R.~Platt, J.-W. van~de Meent, and L.~L. Wong.
\newblock Action priors for large action spaces in robotics.
\newblock \emph{arXiv preprint arXiv:2101.04178}, 2021.

\bibitem[Gu et~al.(2017)Gu, Holly, Lillicrap, and Levine]{gu2017deep}
S.~Gu, E.~Holly, T.~Lillicrap, and S.~Levine.
\newblock Deep reinforcement learning for robotic manipulation with
  asynchronous off-policy updates.
\newblock In \emph{2017 IEEE international conference on robotics and
  automation (ICRA)}, pages 3389--3396. IEEE, 2017.

\bibitem[Singh et~al.(2019)Singh, Yang, Hartikainen, Finn, and
  Levine]{singh2019end}
A.~Singh, L.~Yang, K.~Hartikainen, C.~Finn, and S.~Levine.
\newblock End-to-end robotic reinforcement learning without reward engineering.
\newblock \emph{arXiv preprint arXiv:1904.07854}, 2019.

\bibitem[Zeng et~al.(2018)Zeng, Song, Welker, Lee, Rodriguez, and
  Funkhouser]{zeng2018learning}
A.~Zeng, S.~Song, S.~Welker, J.~Lee, A.~Rodriguez, and T.~Funkhouser.
\newblock Learning synergies between pushing and grasping with self-supervised
  deep reinforcement learning.
\newblock In \emph{2018 IEEE/RSJ International Conference on Intelligent Robots
  and Systems (IROS)}, pages 4238--4245. IEEE, 2018.

\bibitem[Zeng et~al.(2020)Zeng, Song, Lee, Rodriguez, and
  Funkhouser]{zeng2020tossingbot}
A.~Zeng, S.~Song, J.~Lee, A.~Rodriguez, and T.~Funkhouser.
\newblock Tossingbot: Learning to throw arbitrary objects with residual
  physics.
\newblock \emph{IEEE Transactions on Robotics}, 36\penalty0 (4):\penalty0
  1307--1319, 2020.

\bibitem[Zhan et~al.(2020)Zhan, Zhao, Pinto, Abbeel, and Laskin]{ferm}
A.~Zhan, P.~Zhao, L.~Pinto, P.~Abbeel, and M.~Laskin.
\newblock A framework for efficient robotic manipulation.
\newblock \emph{arXiv preprint arXiv:2012.07975}, 2020.

\bibitem[Haarnoja et~al.(2018)Haarnoja, Zhou, Abbeel, and Levine]{sac}
T.~Haarnoja, A.~Zhou, P.~Abbeel, and S.~Levine.
\newblock Soft actor-critic: Off-policy maximum entropy deep reinforcement
  learning with a stochastic actor.
\newblock In \emph{International conference on machine learning}, pages
  1861--1870. PMLR, 2018.

\bibitem[Oord et~al.(2018)Oord, Li, and Vinyals]{oord2018representation}
A.~v.~d. Oord, Y.~Li, and O.~Vinyals.
\newblock Representation learning with contrastive predictive coding.
\newblock \emph{arXiv preprint arXiv:1807.03748}, 2018.

\bibitem[Laskin et~al.(2020)Laskin, Srinivas, and Abbeel]{curl}
M.~Laskin, A.~Srinivas, and P.~Abbeel.
\newblock Curl: Contrastive unsupervised representations for reinforcement
  learning.
\newblock In \emph{International Conference on Machine Learning}, pages
  5639--5650. PMLR, 2020.

\bibitem[Coumans and Bai(2016)]{pybullet}
E.~Coumans and Y.~Bai.
\newblock Pybullet, a python module for physics simulation for games, robotics
  and machine learning.
\newblock \emph{GitHub repository}, 2016.

\bibitem[Laskin et~al.(2020)Laskin, Lee, Stooke, Pinto, Abbeel, and
  Srinivas]{rad}
M.~Laskin, K.~Lee, A.~Stooke, L.~Pinto, P.~Abbeel, and A.~Srinivas.
\newblock Reinforcement learning with augmented data.
\newblock \emph{arXiv preprint arXiv:2004.14990}, 2020.

\bibitem[Paszke et~al.(2017)Paszke, Gross, Chintala, Chanan, Yang, DeVito, Lin,
  Desmaison, Antiga, and Lerer]{pytorch}
A.~Paszke, S.~Gross, S.~Chintala, G.~Chanan, E.~Yang, Z.~DeVito, Z.~Lin,
  A.~Desmaison, L.~Antiga, and A.~Lerer.
\newblock Automatic differentiation in {PyTorch}.
\newblock In \emph{NIPS Autodiff Workshop}, 2017.

\bibitem[Wohlkinger et~al.(2012)Wohlkinger, Aldoma, Rusu, and Vincze]{3dnet}
W.~Wohlkinger, A.~Aldoma, R.~B. Rusu, and M.~Vincze.
\newblock 3dnet: Large-scale object class recognition from cad models.
\newblock In \emph{2012 IEEE international conference on robotics and
  automation}, pages 5384--5391. IEEE, 2012.

\bibitem[Kingma and Ba(2014)]{adam}
D.~P. Kingma and J.~Ba.
\newblock Adam: A method for stochastic optimization.
\newblock \emph{arXiv preprint arXiv:1412.6980}, 2014.

\bibitem[Kostrikov et~al.(2020)Kostrikov, Yarats, and Fergus]{drq}
I.~Kostrikov, D.~Yarats, and R.~Fergus.
\newblock Image augmentation is all you need: Regularizing deep reinforcement
  learning from pixels.
\newblock \emph{arXiv preprint arXiv:2004.13649}, 2020.

\end{thebibliography}

\clearpage

\appendix

\section{Network Architecture}
\label{app:network}

\begin{figure*}[t]
    \centering
    \includegraphics[width=0.7\textwidth]{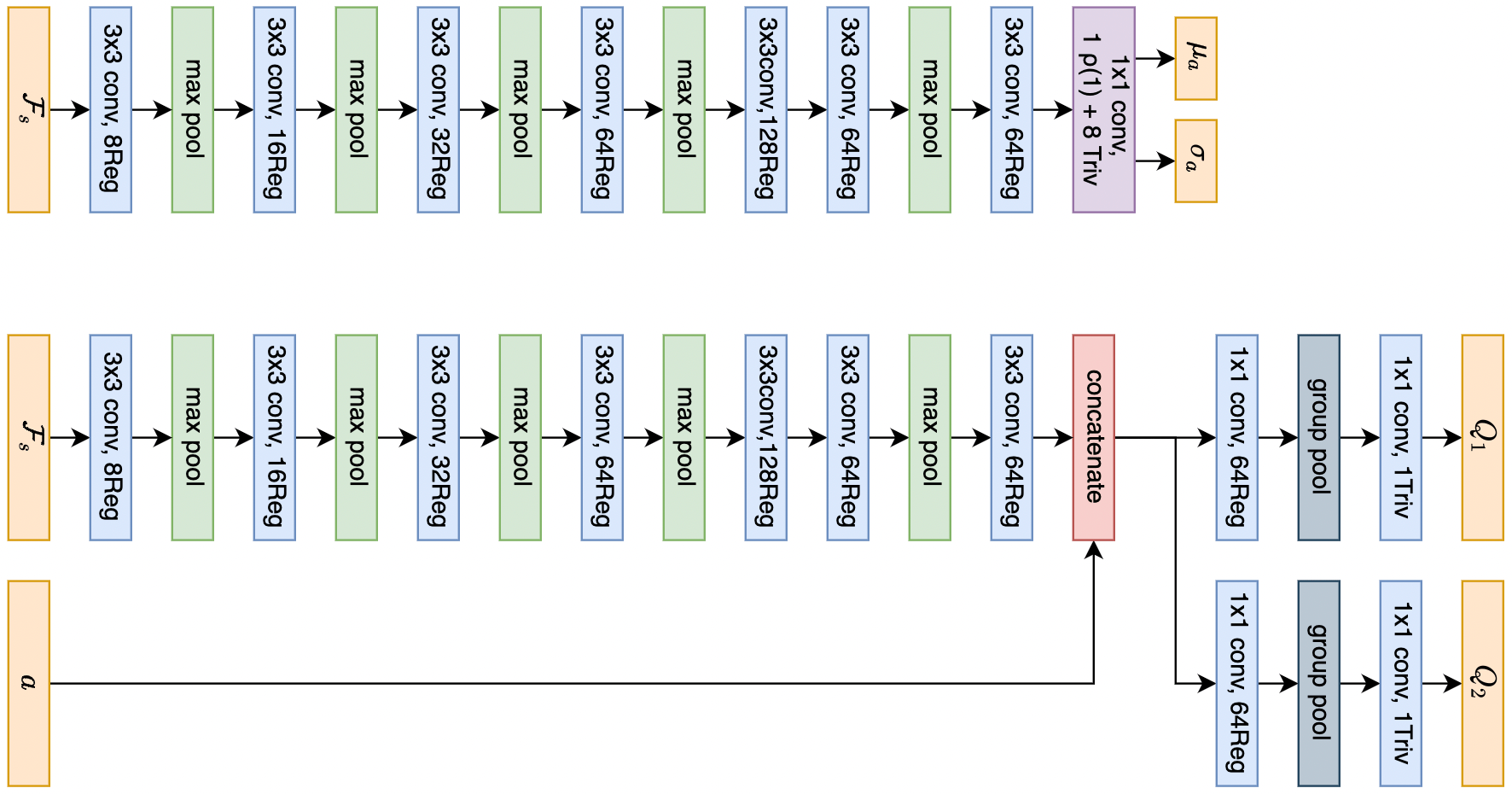}
    \caption{The network architecture of Equivariant SAC. Top: the architecture of the actor network. Bottom: the architecture of the critic network. ReLU non-linearity is omitted in the figure.}
    \label{fig:network}
\end{figure*}

Our equivariant neural networks are implemented using the E2CNN~\cite{e2cnn} library in PyTorch~\cite{pytorch}. The network architecture of the discrete group variations ($C_4, C_8, D_4, D_8$) is illustrated in Figure~\ref{fig:network}. The actor network $\pi$ (Figure~\ref{fig:network} top) takes in $\mathcal{F}_s$ as a trivial representation feature map. The hidden layers are all implemented using the regular representation. The output of $\pi$ is a mixed representation feature map with $1\times1$ spatial dimension (so that it can also be viewed as a vector). This vector consists of 1 $\rho_1$ feature, representing the means of the action component $(x, y)$, and 8 trivial representation features, representing the means of the action component $(z, \theta, \lambda)$ as well as the standard deviations of all action dimensions. The critic network $q$ (Figure~\ref{fig:network} bottom) takes in $\mathcal{F}_s$ as a trivial representation feature map. The upper path of $q$ generates a 64-channel regular representation feature map with $1\times 1$ spatial dimensions. This regular representation feature map is concatenated with the action $a$, which is encoded as 1 $\rho_1$ feature (representing $(x, y)$) and 3 $\rho_0$ features (representing $(z, \theta, \lambda)$). The concatenated feature map is sent to two separate blocks to generate two $Q$ estimates in the form of $1\times 1$ trivial representation feature map. In both the actor and critic network, we use ReLU as the activation function.

For the continuous group variations ($\SO(2), \mathrm{O}_2$), we define the networks with a maximum frequency of 3 (since the continuous group $\SO(2), \mathrm{O}_2$ have infinite number of irreducible representations, the maximum frequency specifies the maximum frequency of the irreducible representations to build~\cite{e2cnn}). The forms of the input and output feature types are the same as the discrete group variations. However, for the hidden layers, we use $\rho_0^k \oplus \rho_1^k \oplus \rho_2^k \oplus \rho_3^k$ as the representation type for the $\SO(2)$ network. In principle, $\rho_m(g)=\rho_1(mg)$ describes the rotation symmetries in different frequencies. We use $\rho_0^k \oplus \rho_{(1, 0)}^k \oplus \rho_{(1, 1)}^k \oplus \rho_{(1, 2)}^k \oplus \rho_{(1, 3)}^k$ for the hidden layers in the $\mathrm{O}(2)$ network. The group $\mathrm{O}(2)$ is generated by rotations $\mathrm{Rot}(\theta)$ and a reflection $f$ over the $x$-axis. For $k > 0$, $\rho_{(1,k)}$ is the irreducible representation of $\mathrm{O}(2)$ on $\mathbb{R}^2$ in which rotations $\mathrm{Rot}(\theta)$ rotate vectors about the origin by $k \theta$ and $f$ reflects vectors over the $x$-axis. We use the gated nonlinearity as the activation function. 

\section{Simulation Environment Details}
\label{app:sim_env}

In all simulation environments, the workspace has a size of $0.3m\times 0.3m\times 0.24m$. All workspace is inside a bin with a bounding box size of $0.45m\times 0.45m\times 0.1m$. The bottom size of the bin is $0.3m\times 0.3m$. We use a simulated Franka Panda arm in all simulations. All environments raise a sparse reward, i.e., +1 when the agent finishes the task, and 0 otherwise. The maximum time steps for all tasks is 50. At the beginning of each episode, the gripper is moved to the center of the workspace.

\subsection{Block Picking} 
In the Block Picking environment, the robot needs to pick up a block with a size of $5cm\times 5cm\times 5cm$. At the start of each episode, the block is initialized with random position and orientation. The goal is to grasp the block and lift it s.t. the gripper is at least $0.15m$ above the ground. The observation in this environment covers an area of $0.3m\times 0.3m$.

\begin{figure*}[t]
    \centering
    \includegraphics[width=0.7\textwidth]{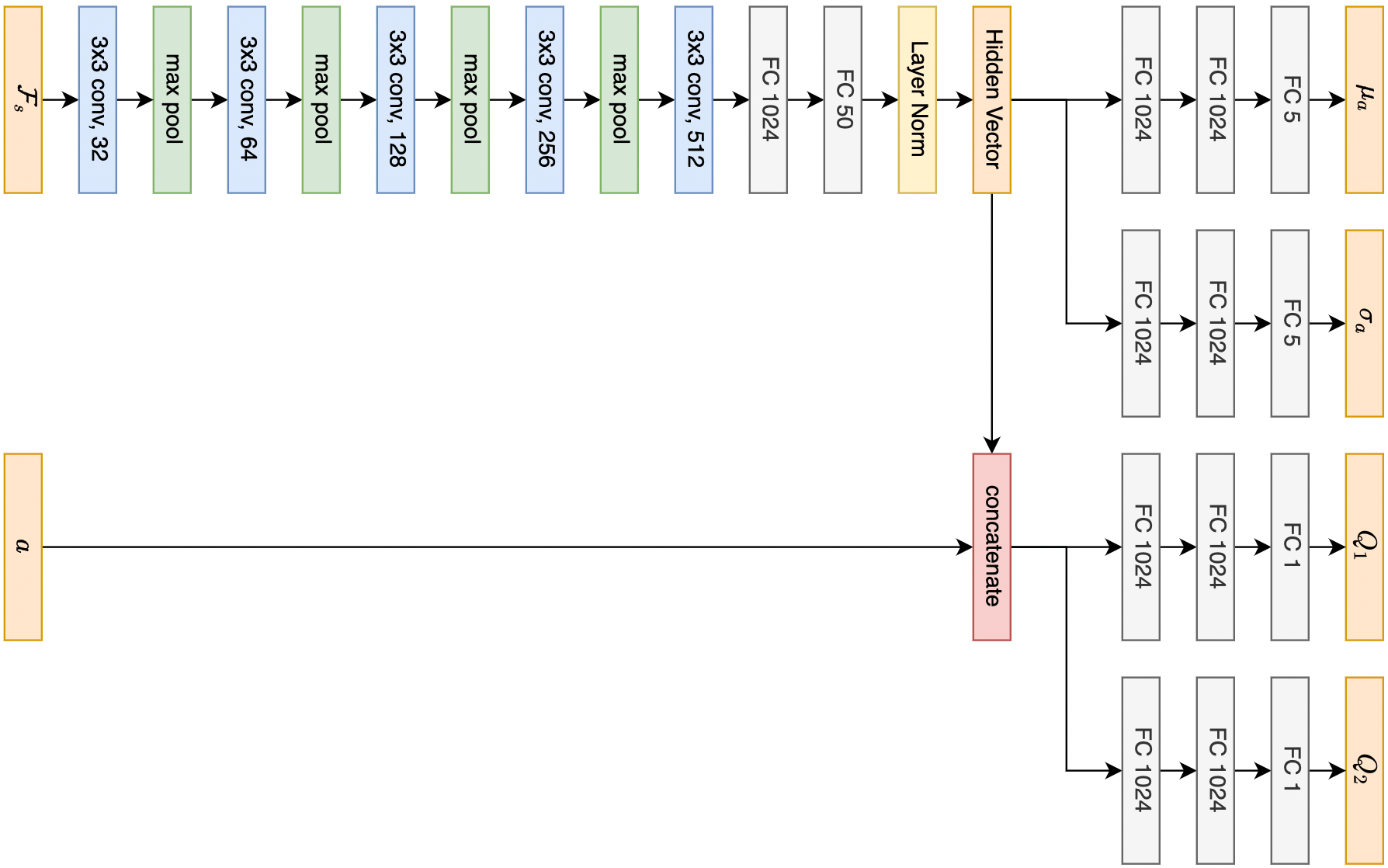}
    \caption{The network architecture of the FERM baseline. ReLU non-linearity is omitted in the figure.}
    \label{fig:ferm_network}
\end{figure*}

\subsection{Clutter Grasping}
In the Clutter Grasping environment, the robot needs to pick up an object from a clutter of at most five objects. At the start of training, five random objects are initialized with random position and orientation. The shapes of the objects are randomly sampled from the object set shown in Figure~\ref{fig:grasp_obj_set}. The object set contains 76 objects derived from the 3DNet~\cite{3dnet} dataset. Every time the agent successfully grasps all five objects, the environment will re-generate five random objects with random positions and orientations. If an episode terminates with any remaining objects in the bin, the object will not be re-initialized. The goal of this task is to grasp any object and lift it s.t. the gripper is at least $0.15m$ above the ground. The observation in this environment covers an area of $0.3m\times 0.3m$.

\subsection{Block Pushing}
In the Block Pushing environment, the robot needs to push a block with a size of $5cm\times 5cm\times 5cm$ to the goal area with a size of $9cm\times 9cm$. At the start of each episode, the block is initialized with a random position and orientation, and the goal is initialized with a random position at least $9cm$ away from the block. The goal is to push the block s.t. the distance between the block's center and the goal's center is within $5cm$. The observation in this environment covers an area of $0.45m\times 0.45m$. The goal area is drawn on the observation by adding $2cm$ to the height values.

\subsection{Block in Bowl}
In the Block in Bowl environment, the robot needs to pick up a block with a size of $5cm\times 5cm\times 5cm$ and place it inside a bowl with a bounding box size of $16cm\times16cm\times7cm$. At the start of each episode, the block and the bowl are initialized with a random position and orientation. The observation in this environment covers an area of $0.45m\times 0.45m$.

\section{On-Robot Learning Details}
\label{app:real_env}
In the on-robot learning environment, the two bins have a bottom size of $0.25m\times 0.25m$. The bounding box of each bin has a size of $0.4m\times 0.4m\times 0.11m$. The workspace of the arm is inside one of the two bins, with a size of $0.23m\times 0.23m\times 0.2m$.
Similar to simulation environments, all real-world environments raise a sparse reward, +1 when success and 0 otherwise. We implement a collision protection algorithm to prevent the arm from colliding with the environment by reading the force applied to the end-effector. A -0.1 reward is given when a protective stop is triggered on the UR5. The maximum number of time steps per episode for all environments is 50. Reset and reward functions are implemented specifically for different tasks as follows.

\subsection{Block picking}
In the block picking environment, only one bin will be used throughout the training. The block has a size of $5cm\times 5cm\times 5cm$. A reward will be given when the block is grasped and lifted for 0.1m above the ground by reading the signals from the gripper. At the start of each episode, the block is reset to a random position and orientation in the current bin. The observation covers an area of $0.3m\times 0.3m$.

\begin{figure}[t]
\centering
\subfloat[]{
\includegraphics[height=0.35\linewidth]{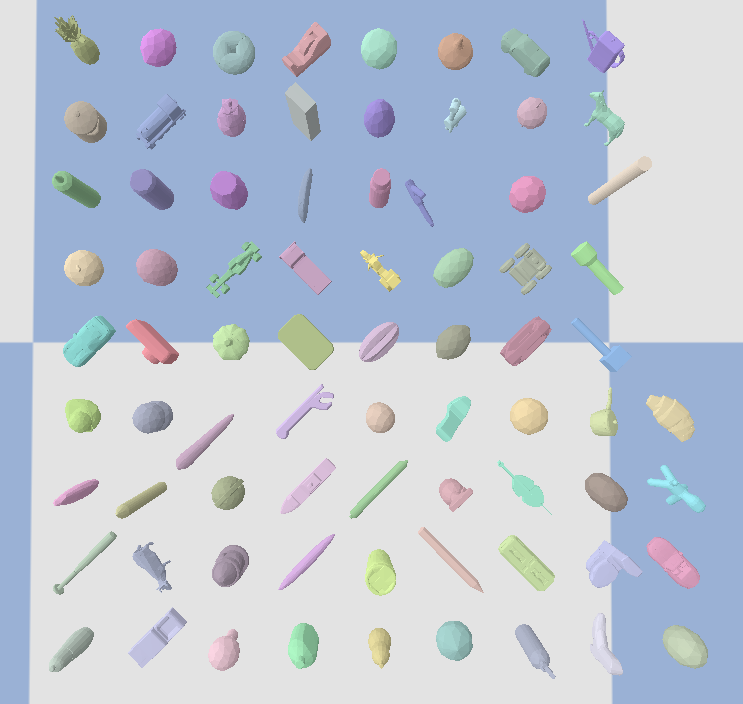}
\label{fig:grasp_obj_set}
}
\subfloat[]{
\includegraphics[height=0.35\linewidth]{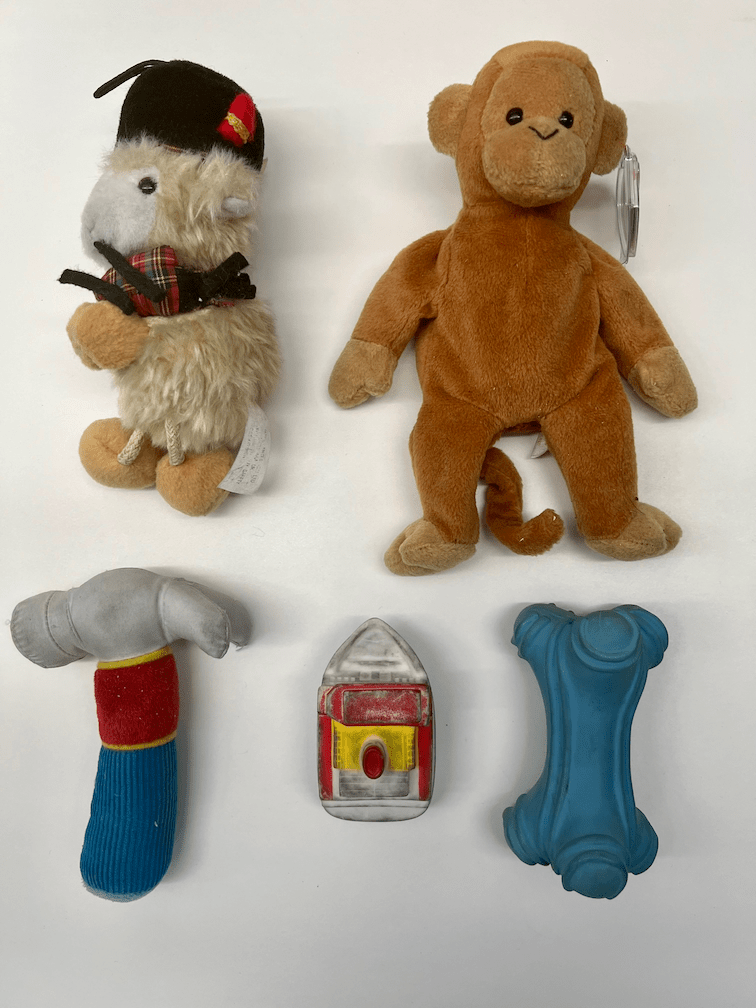}
\label{fig:on_robot_grasp_set}
}
\caption{(a) The object set for Clutter Grasping in simulation contains 76 objects from the 3DNet~\cite{3dnet} dataset. (b) The object set for on-robot Clutter Grasping}
\end{figure}



\subsubsection{Clutter Grasping}

\begin{figure}[t]
\centering
\subfloat[]{\includegraphics[width=0.48\linewidth]{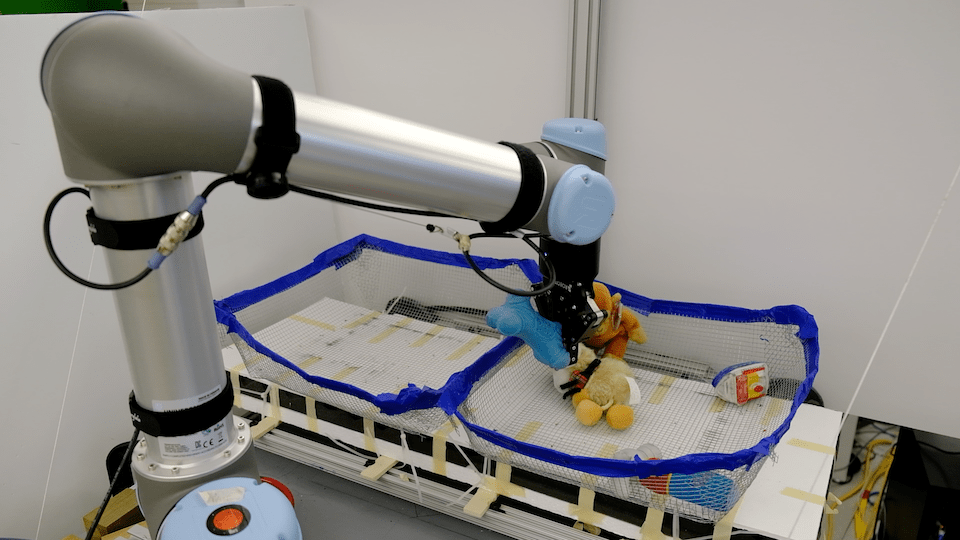}
}
\subfloat[]{\includegraphics[width=0.48\linewidth]{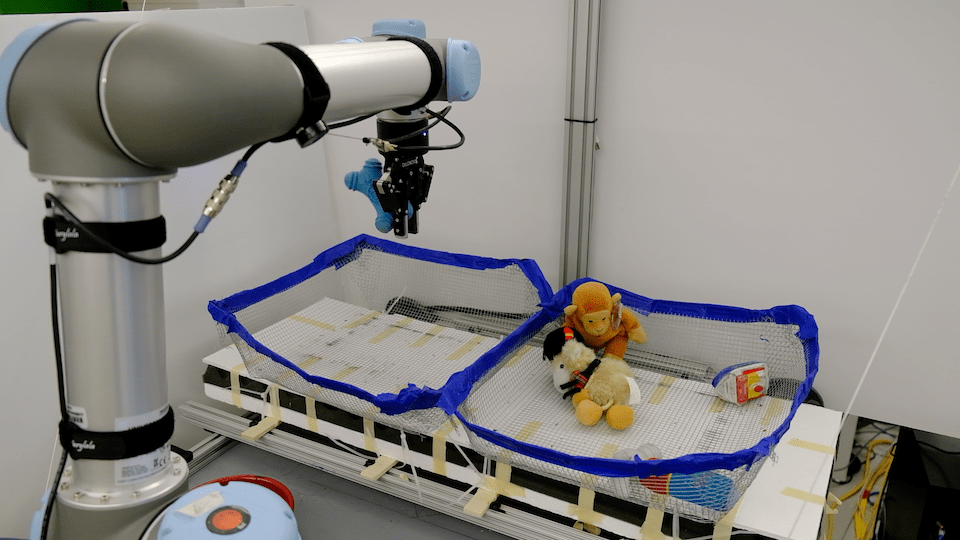}
}\\
\subfloat[]{\includegraphics[width=0.48\linewidth]{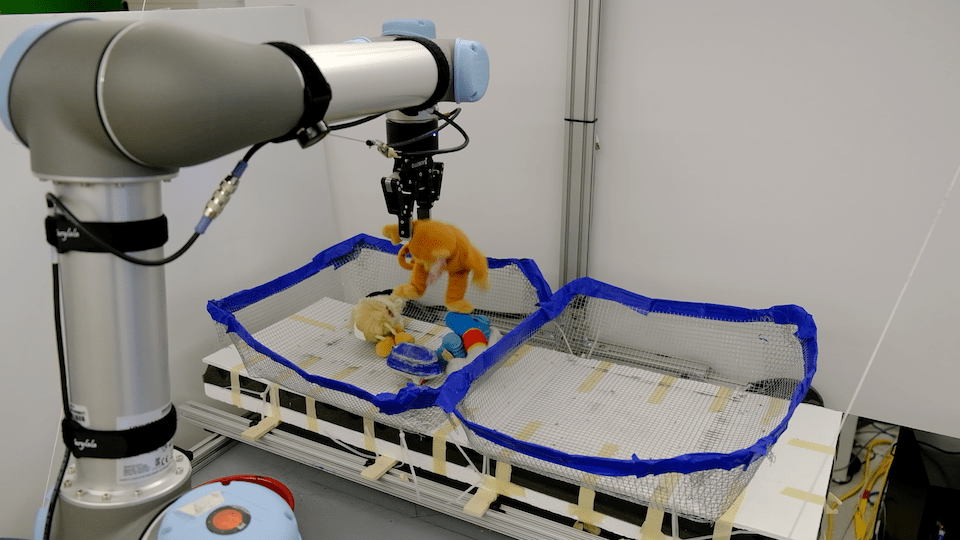}
}
\subfloat[]{\includegraphics[width=0.48\linewidth]{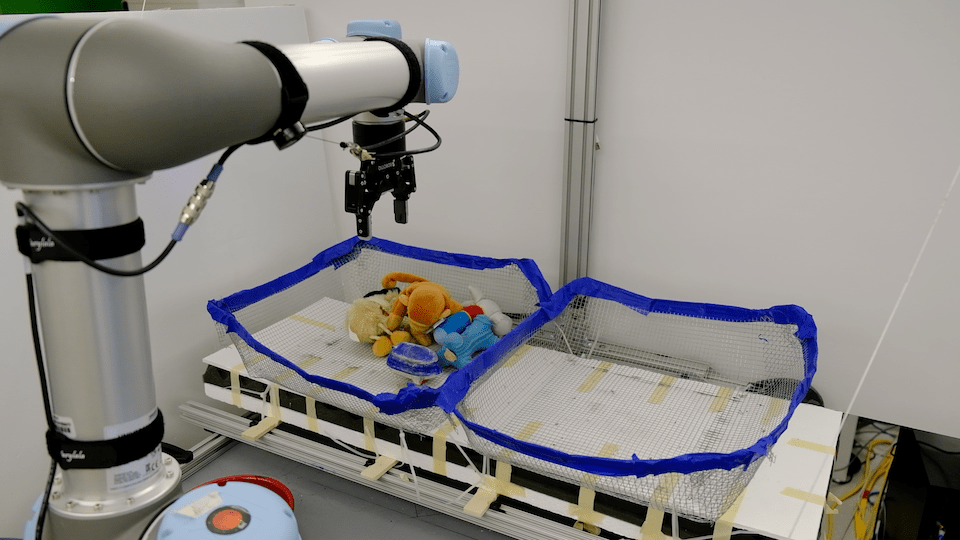}
}
\caption{The reset process of Clutter Grasping. (a) The robot successfully grasps an object in the active workspace (right bin). (b) The robot places the grasped object in the reset bin (left bin). (c) The robot grasps and places the last object from the active workspace (right bin) to the reset bin (left bin). (d) The robot switches the active workspace to the left bin.}
\label{fig:grasp_reset}
\end{figure}

The Clutter Grasping environment contains a clutter of five objects, as is shown in Figure~\ref{fig:on_robot_grasp_set}. In this experiment, one bin is used as the active workspace, and the other bin is used as the reset bin. The goal is to grasp an object and lift it to 0.1m above the ground. Once grasped, the object will be reset with random positions and orientations into the reset bin. After all objects in the active workspace are grasped, the active workspace and the reset bin is swapped. This process is shown in Figure~\ref{fig:grasp_reset}. If the robot fails to grasp any object for five consecutive episodes, all remaining objects in the active workspace will be moved to the reset bin. The observation covers an area of $0.3m\times 0.3m$.

\subsection{Block Pushing}
In the Block Pushing environment, only one bin will be used throughout the training. The goal is to move the block s.t. more than 77 pixels of the block are within the goal area. The block has a size of $5cm\times 5cm\times 5cm$. At the beginning of each episode, a goal area of $0.77m\times 0.77m$ is virtually generated at least $9cm$ away from the block. The goal area is drawn to the observation by adding the height value by $0.02m$. When resetting, the arm will move the block to a new random position. The observation covers an area of $0.45m\times 0.45m$, in order to get a larger field of view that includes both object and goal area.

\begin{figure}[t]
\centering
\subfloat[]{\includegraphics[width=0.48\linewidth]{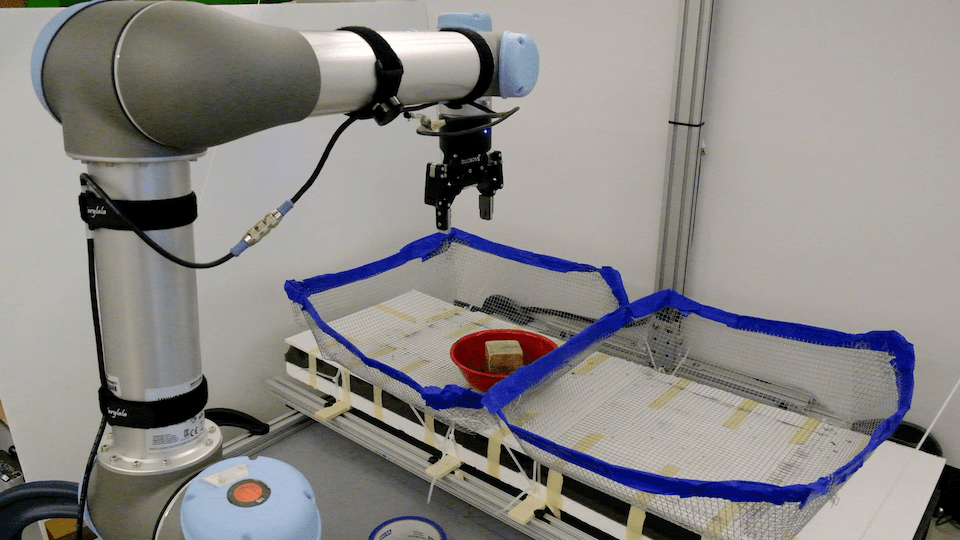}
}
\subfloat[]{\includegraphics[width=0.48\linewidth]{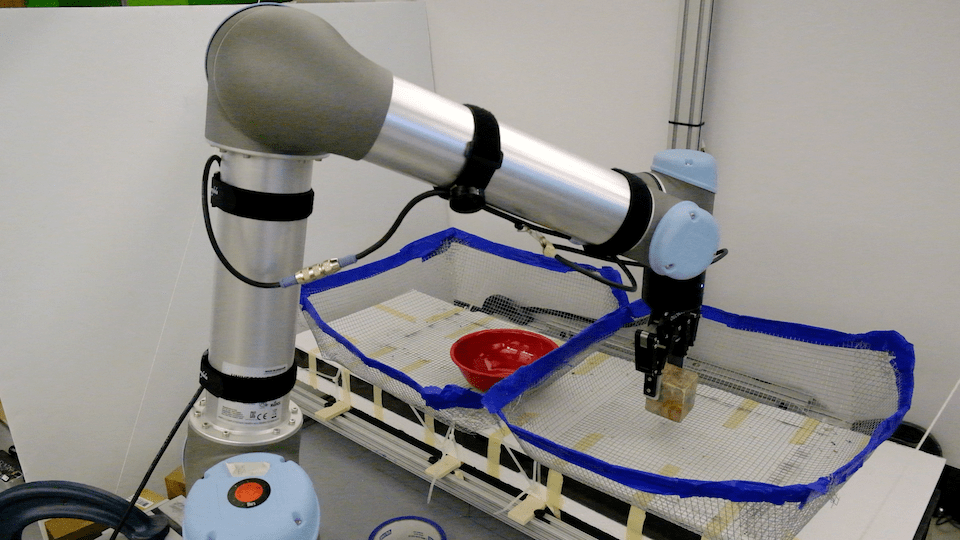}
}\\
\subfloat[]{\includegraphics[width=0.48\linewidth]{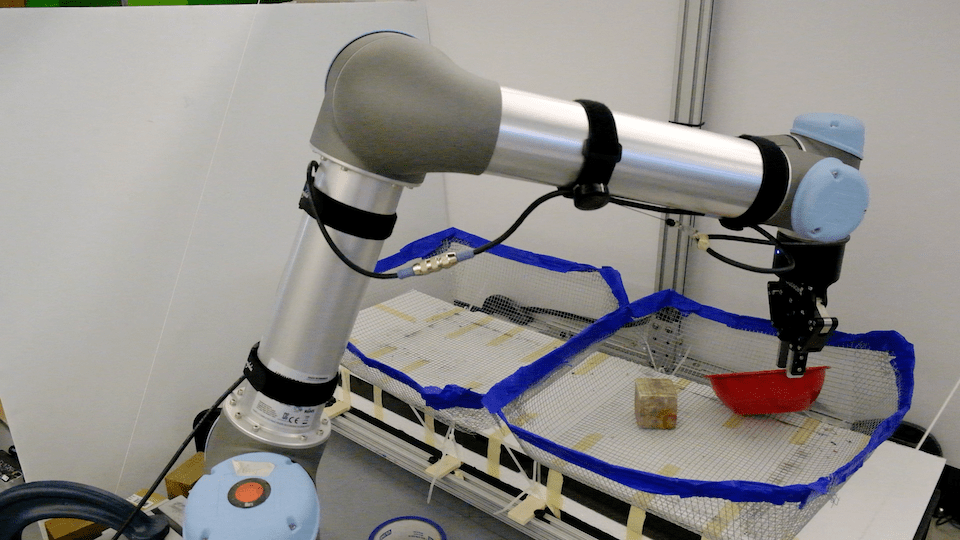}
}
\subfloat[]{\includegraphics[width=0.48\linewidth]{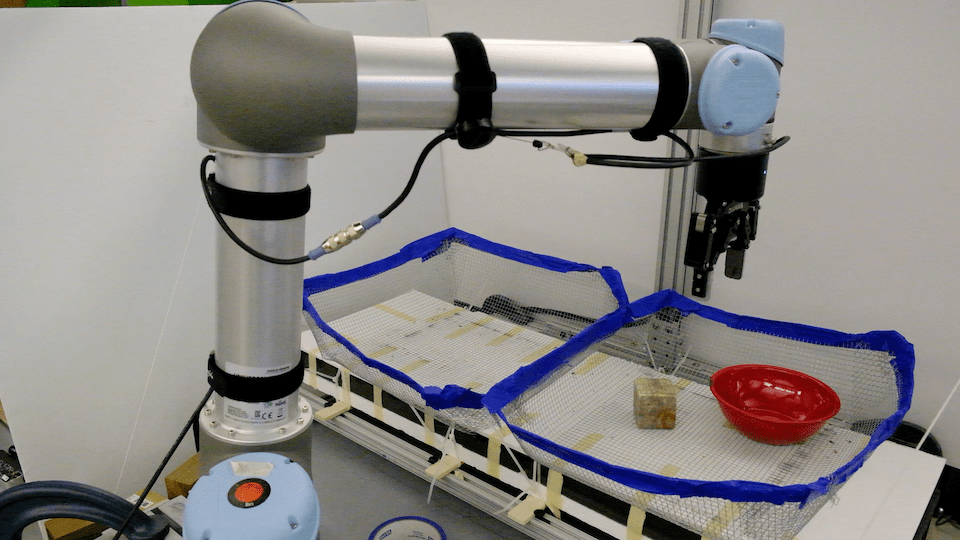}
}
\caption{The reset process of Block in Bowl. (a) The robot finishes an episode in the active workspace (left bin). (b) The robot picks up the block and places it in the reset bin (right bin). (c) The robot picks up the bowl and places it in the reset bin (right bin). (d) The robot switches the active workspace to the right bin.}
\label{fig:bowl_reset}
\end{figure}

\subsection{Block in Bowl}
In the Block in Bowl environment, we use one bin as the active workspace and the other bin as the reset bin. The goal is to grasp the block and then put it into the bowl. The bowl has a radius of $8cm$ and the block has a size of $5cm\times 5cm\times 5cm$. At the beginning of each episode, both the block and the bowl will be reset into the reset bin with random positions and orientations, keeping at least $14cm$ away from each other. The active workspace and the reset bin are then swapped. This process is shown in Figure~\ref{fig:bowl_reset}. The reset function utilizes circle detection for grasping the bowl. The observation covers an area of $0.45m\times 0.45m$ to provide a larger view to cover both the block and the bowl.

\section{Training Details}
\label{app:traininig_detail}
The pixel size of the observation is $128\times 128$ for all methods except for the FERM baseline and RAD crop baseline, where the observation's pixel size is $142\times 142$ and will be cropped to $128\times 128$. 

We train the networks using the Adam~\cite{adam} optimizer with a learning rate of $10^{-3}$. We perform one SGD step per environmental step. The entropy temperature $\alpha$ of SAC is initialized at $10^{-2}$. The target entropy is $-5$. The discount factor is $\gamma=0.99$. We use a target network for the critic network and soft target update with $\tau=10^{-2}$. The replay buffer has a capacity of 100,000 transitions. The mini-batch size is 64.

\section{Baseline Details}
\label{app:baseline_detail}

The FERM~\citep{ferm} baselines use random crop for data augmentation. The random crop crops a $142\times 142$ state image to the size of $128\times 128$. As in ~\cite{ferm}, the contrastive encoder has an output size of 50, and the contrastive encoder is pre-trained for 1.6k steps using the expert data. Figure~\ref{fig:ferm_network} shows the network architecture for our baseline FERM.

\section{Buffer Augmentation}
\label{appendix:exp_buffer_aug}
\begin{figure}[t]
\centering
\includegraphics[width=\textwidth]{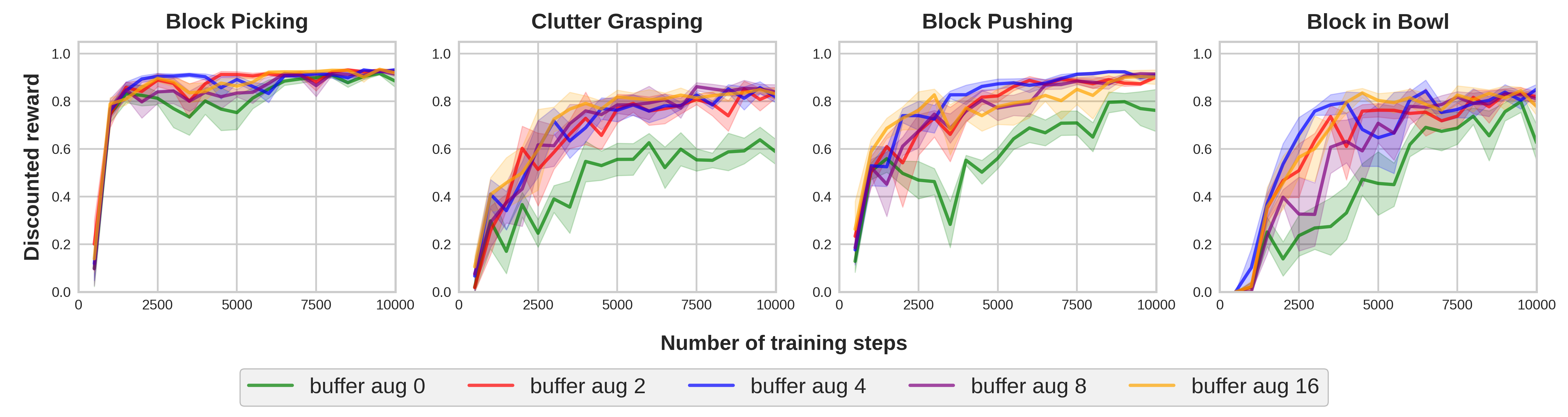}
\caption{Comparison of Equivariant SAC equipped with rotational data augmentation in the replay buffer. \plotDescription}
\label{fig:exp_buffer_aug}
\end{figure}

In this ablation study, we consider five different numbers of augmentations for each transition: 0, 2, 4, 8, and 16 (0 means no extra augmented transition is added). Figure~\ref{fig:exp_buffer_aug} shows the result. First, note that no augmentation at all (green) is always the worst-performing variation, suggesting that providing extra augmented samples to the agent is beneficial. Second, note that more augmentation does not necessarily mean better performance (e.g., buffer aug 8 (purple) and buffer aug 16 (orange) underperforms buffer aug 4 (blue) in Block in Bowl). Four augmentations (blue) shows the best performance overall.

\section{Effect of Expert Demonstration}
\label{appendix:expert}
Expert demonstrations are critical when solving challenging sparse rewards tasks. Without it, the agent must search randomly for a goal state and this can take a long time. 
This section evaluates two essential factors in injecting expert demonstration to Equivariant SAC: the number of expert demonstrations needed and if a behavior cloning loss will be beneficial.

\subsection{Number of Expert Demonstration}
\begin{figure}[t]
\centering
\includegraphics[width=\textwidth]{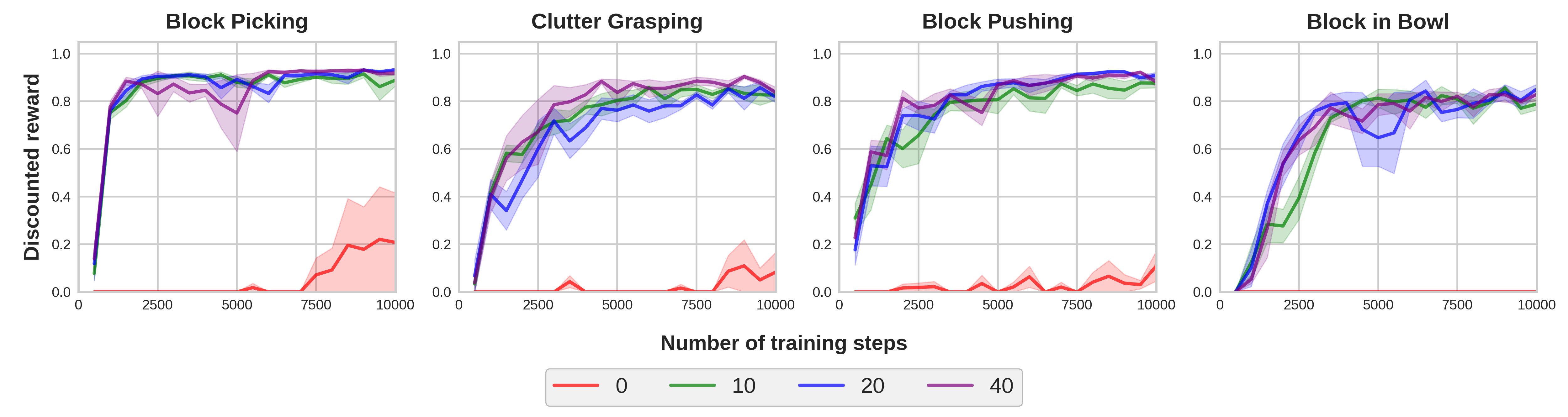}
\caption{Comparison of Equivariant SAC with different amount of expert demonstration episodes. \plotDescription}
\label{fig:exp_n_expert}
\end{figure}

This experiment studies two questions: 1) if expert demonstration is necessary when using Equivariant SAC to solve our tasks; 2) if it is necessary, how many demonstrations are needed. We consider four different amount of expert demonstration episodes provided to the agent: 0, 10, 20, and 40. Figure~\ref{fig:exp_n_expert} shows the comparison result. First, note that expert demonstration is always required since the variation without any demonstration (red) struggles to learn a good policy in all four environments. Second, we found that the Equivariant SAC can do well with just 10 or 20 expert demonstrations.
The fact that we require so few demonstrations is especially important in situations where it is a human who must provide the demonstrations.


\subsection{SACfD}
\label{app:sacfd}
\begin{figure}[t]
\centering
\includegraphics[width=\textwidth]{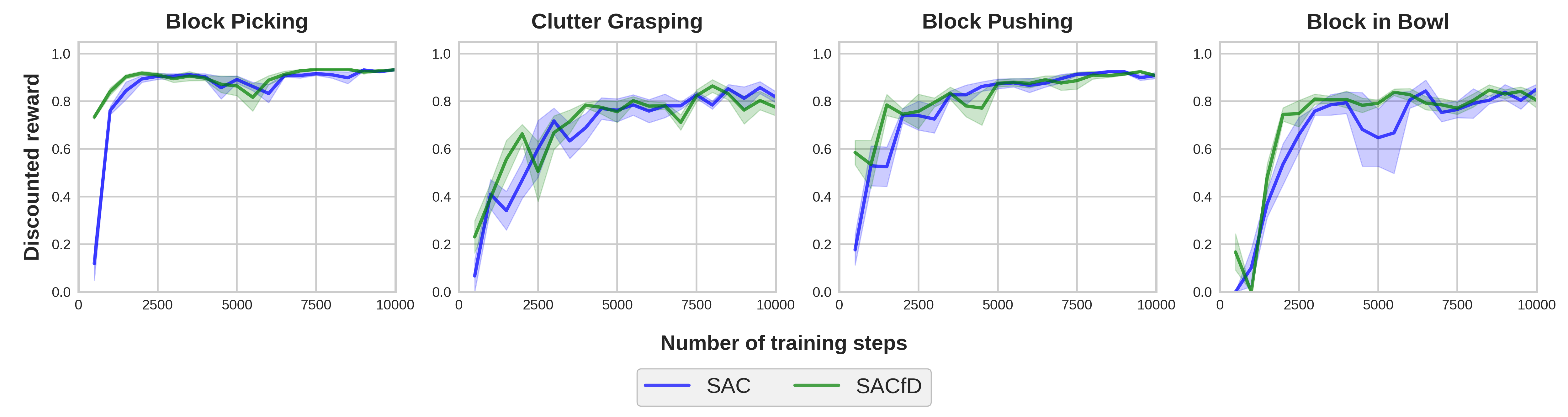}
\caption{Comparison of Equivariant SAC with Equivariant SAC from Demonstration (SACfD). \plotDescription}
\label{fig:exp_sacfd}
\end{figure}

In this experiment, we evaluate if an auxiliary demonstration loss will be beneficial for Equivariant SAC. We use SACfD~\cite{iclr} as the baseline for incorporating a demonstration loss to the actor:
\begin{equation}
\mathcal{L}_{\mathrm{SACfD}}^{\mathrm{actor}}=\mathbbm{1}_{e} \left[ \frac{1}{2}((a\sim \pi(s)) - a_e)^2\right],
\end{equation}
where $\mathbbm{1}_{e}=1$ if the sampled transition is an expert demonstration and 0 otherwise, $a\sim \pi(s)$ is an action sampled from the output Gaussian distribution of $\pi (s)$, and $a_e$ is the expert action. This demonstration loss is used in addition to the loss of SAC to incline its policy to the expert policy. Figure~\ref{fig:exp_sacfd} shows the comparison between Equivariant SACfD (green) and Equivariant SAC (blue). Interestingly, Equivariant SACfD outperforms Equivariant SAC in three of the environments, but underperforms in Clutter Grasping. This is because the expert policy for Clutter Grasping is sub-optimal (since the planner does not have access to the optimal grasping point of each random object), while the expert for the other three tasks are nearly optimal.
The performance of SACfD indicates that when the expert demonstration is sub-optimal, an auxiliary demonstration loss will harm the performance of Equivariant SAC. This is an important finding because the planner that generates the expert policy in on-robot learning is often sub-optimal (since it does not have access to the pose of the objects as in the simulation).

\section{\edit{Comparing Equivariant SAC with CNN baselines}}

\begin{figure}[t]
\centering
\includegraphics[width=\textwidth]{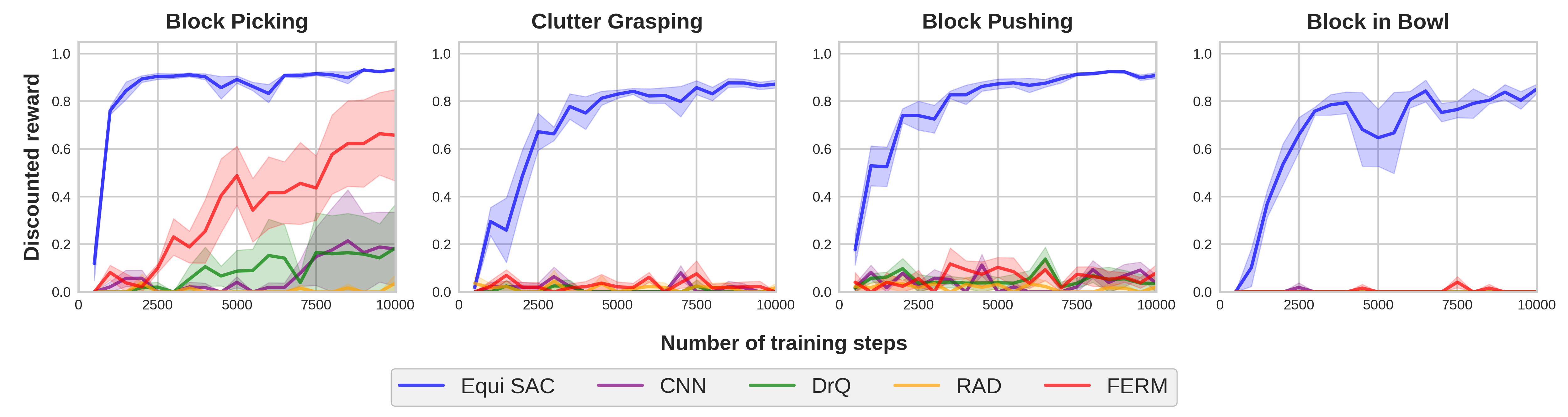}
\caption{\edit{Comparison of Equivariant SAC (blue) with baselines using conventional CNN. \plotDescription}}
\label{fig:exp_cnn_baseline}
\end{figure}

\begin{figure}[t]
\centering
\includegraphics[width=\textwidth]{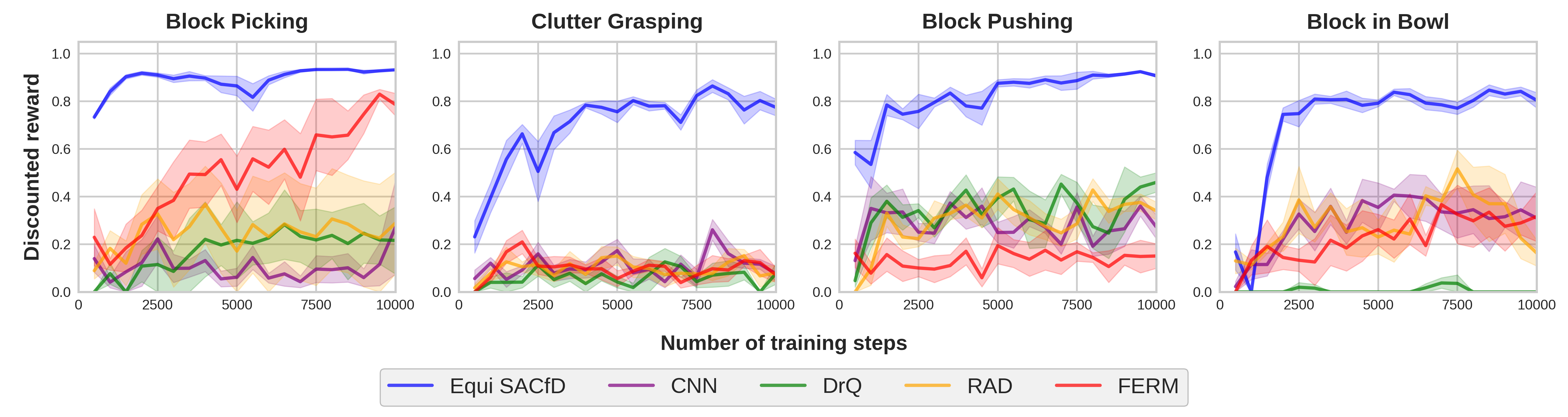}
\caption{\edit{Comparison of Equivariant SACfD (blue) with baselines using conventional CNN. \plotDescription}}
\label{fig:exp_cnn_baseline_sacfd}
\end{figure}

\edit{In this experiment, we compare the Equivariant SAC (Equi SAC) with some sample efficient baselines using standard convolutional neural networks and data augmentation. We consider five baselines: 1) CNN: standard SAC~\cite{sac} implemented using conventional CNN instead of equivariant networks. 2) DrQ~\cite{drq}: 1) with random shift data augmentation. At each training step, both the $Q$-targets and the TD losses are calculated by averaging over two random-shift augmented transitions. 3) RAD~\cite{rad}: 1) with random crop data augmentation. At each training step, each transition in the minibatch is applied with a random crop data augmentation. 4): FERM~\cite{ferm} similar architecture as 1) with an extra contrastive loss term that learns an invariant encoder from random crop augmentation. As is shown in Figure~\ref{fig:exp_cnn_baseline}, Equivariant SAC (blue) significantly outperforms all baselines, where non of the baselines can solve Clutter Grasping, Block Pushing and Block in Bowl.}

\edit{We further perform the same experiment using SACfD (Appendix~\ref{app:sacfd}) instead of SAC. As is shown in Figure~\ref{fig:exp_cnn_baseline_sacfd}. SACfD improves the performance of the CNN baselines, however, they still underperform Equivariant SACfD (blue) dramatically.}

\end{document}